\documentclass[sigconf,nonacm]{acmart}
\usepackage{algorithmicx,algorithm}
\usepackage{algpseudocode}
\usepackage{algorithm,algpseudocode}
\usepackage{makecell,booktabs}
\usepackage{multirow}
\usepackage{multicol}  
\usepackage{bm}
\usepackage{siunitx}
\usepackage{xcolor}
\usepackage{hyperref}
\usepackage{enumitem}
\usepackage{subfigure}
\usepackage{xspace}
\usepackage{cleveref}
\usepackage{etoolbox}
\usepackage{pdfcomment}
\usepackage{amsmath}
\usepackage[normalem]{ulem}
\usepackage{soul}
\hypersetup{hidelinks}

\crefname{section}{Sec.}{Sec.}
\crefname{figure}{Fig.}{Fig.}
\crefname{table}{Table}{Table}
\crefname{algorithm}{Algorithm}{Algorithm}
\crefname{equation}{Eq.}{Eq.}
\crefname{appendix}{\textbf{Appendix}}{\textbf{Appendix}}

\newcommand{\multirowoffset}{-0.5\dimexpr \aboverulesep + \belowrulesep + \cmidrulewidth}

\newcommand{\stkout}[1]{\ifmmode\text{\sout{\ensuremath{#1}}}\else\sout{#1}\fi}
\newcommand{\udline}[1]{\ifmmode\text{\uline{\ensuremath{#1}}}\else\uline{#1}\fi}

\def\model{ControlTraj\xspace}
\begin{document}
\begin{sloppypar}   
\title{ControlTraj: Controllable Trajectory Generation with Topology-Constrained Diffusion Model}


\author{Yuanshao Zhu$^{\ast,1,2,3}$,~James Jianqiao Yu$^{4,\dag}$,~Xiangyu Zhao$^{2,\dag}$, Qidong Liu$^2$, Yongchao Ye$^2$ \\ Wei Chen$^3$, Zijian Zhang$^2$, Xuetao Wei$^1$, Yuxuan Liang$^{3,\dag}$}
\thanks{$^{\ast}$Work was done at Hong Kong University of Science and Technology (Guangzhou)\\
$^{\dag}$Corresponding authors.}
\affiliation{
  \institution{$^1$Southern University of Science and Technology,~~$^2$City University of Hong Kong\\$^3$The Hong Kong University of Science and Technology (Guangzhou),~~$^4$York University}
  \country{}
  \address{}
\text{zhuys2019@mail.sustech.edu.cn,~james.yu@york.ac.uk,~xianzhao@cityu.edu.hk,~liuqidong@stu.xjtu.edu.cn}\\
  \text{yongchao.ye@my.cityu.edu.hk,~onedeanxxx@gmail.com,~zhangzj2114@mails.jlu.edu.cn}
\text{weixt@sustech.edu.cn,~~yuxliang@outlook.com}
}

\renewcommand{\shortauthors}{Yuanshao Zhu et al.}

\begin{abstract}
Generating trajectory data is among promising solutions to addressing privacy concerns, collection costs, and proprietary restrictions usually associated with human mobility analyses.
However, existing trajectory generation methods are still in their infancy due to the inherent diversity and unpredictability of human activities, grappling with issues such as fidelity, flexibility, and generalizability.
To overcome these obstacles, we propose \textbf{\model}, a \textbf{Control}lable \textbf{Traj}ectory generation framework with the topology-constrained diffusion model.
Distinct from prior approaches, \model utilizes a diffusion model to generate high-fidelity trajectories while integrating the structural constraints of road network topology to guide the geographical outcomes.
Specifically, we develop a novel road segment autoencoder to extract fine-grained road segment embedding.
The encoded features, along with trip attributes, are subsequently merged into the proposed geographic denoising UNet architecture, named GeoUNet, to synthesize geographic trajectories from white noise.
Through experimentation across three real-world data settings, \model demonstrates its ability to produce human-directed, high-fidelity trajectory generation with adaptability to unexplored geographical contexts.
\end{abstract}




\maketitle

\section{Introduction}
The proliferation of GPS technology has revolutionized our understanding of human mobility patterns, yielding profound implications for urban planning \cite{ruan2020learning}, location-based services \cite{wang2021survey}, and beyond \cite{trajode,wang2021trajectory}.
However, acquiring real-world GPS trajectory data frequently confronts formidable obstacles, including privacy concerns \cite{zhu2021semi}, data collection costs \cite{zheng2015trajectory}, proprietary restrictions \cite{cao2021generating,fang2021dragoon}, and a myriad of regulatory barriers \cite{chen2024deep,xu2022metaptp}.
These challenges substantially hinder researchers' ability to access data that faithfully captures the complexity of human mobility.
In this context, trajectory generation emerges as a compelling solution to the problem, offering the potential to create synthetic yet realistic GPS trajectories to bypass these limitations \cite{zhu2023difftraj}.
By generating synthesized trajectories, this solution can propel research without real datasets and adhere to the stringent privacy and data ownership requirements, thereby broadening research and applications in a series of fields \cite{zheng2014urban,dai2015personalized,guo2018learning,jensen2022digitalization}.

\begin{figure}[t]
    \includegraphics[width=0.95\linewidth]{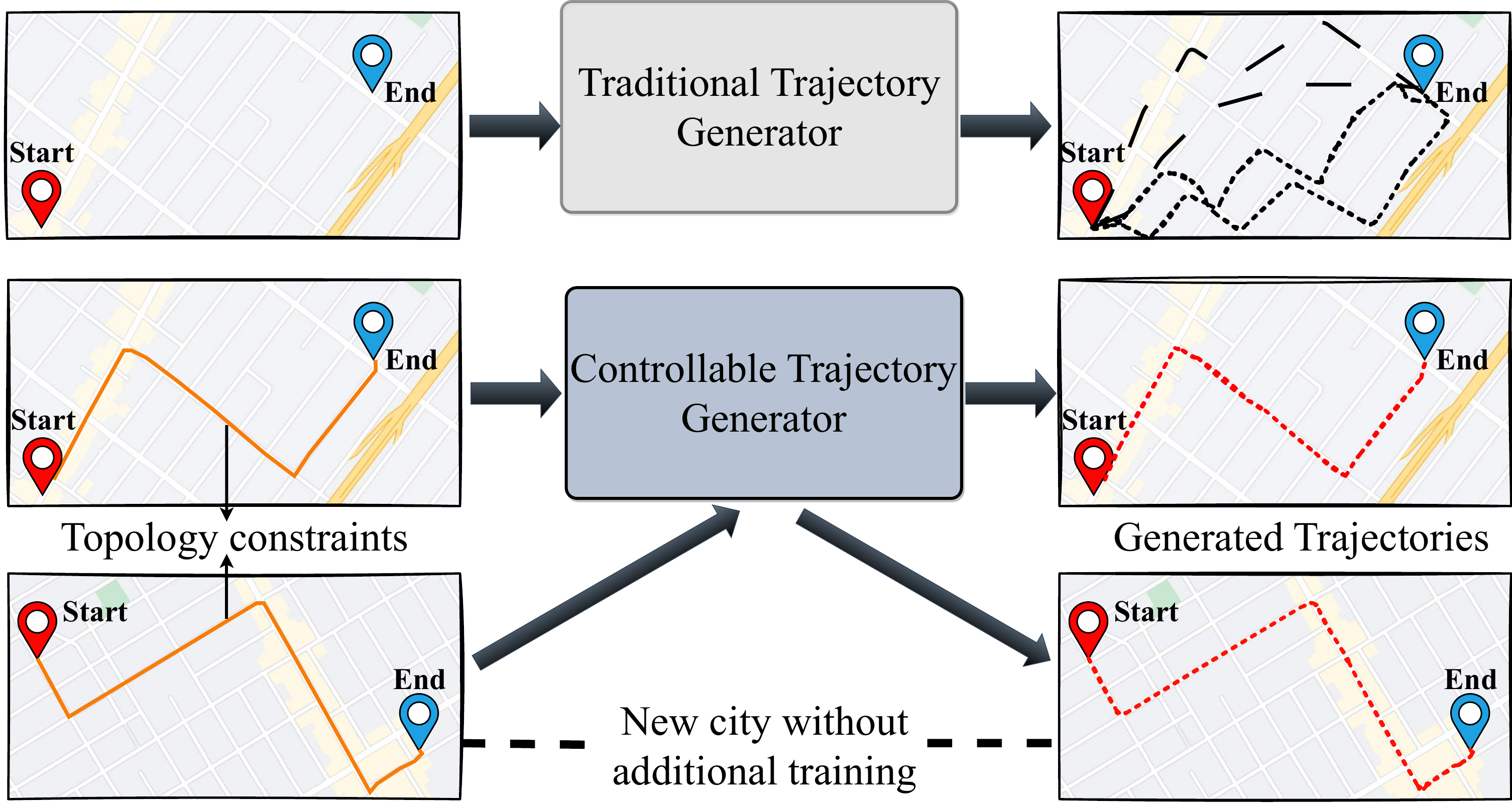}
    \caption{Comparison of trajectory generation models. Our controllable trajectory generator can provide high-fidelity trajectories and adhere to given conditions. Also, it adapts to new urban environments without retraining.}
    \label{fig:intro}
    \vspace*{-6mm}
\end{figure}

To generate trajectories that closely reflect the complexities of real-world scenarios, researchers have pursued various strategies. 
Early attempts, grounded in rule-based or statistical models, managed to offer a degree of interpretability but often fell short in mirroring the stochastic behavior of human mobility \cite{simini2021deep, barbosa2018human}. 
Advancements in machine learning, particularly through the use of Generative Adversarial Networks (GANs) and Variational Autoencoders (VAEs) \cite{Xia2018deeprailway,henke2023condtraj,rao2020lstm}, have led to more sophisticated modeling of complex trajectory distributions. 
Nevertheless, these approaches typically resort to transforming trajectories into simplified data formats like grids \cite{ouyang2018non, yuan2022activity, feng2020learning} or images \cite{cao2021generating, wang2021large}, compromising the fidelity and granularity of the resulting data \cite{luca2021survey}. 
Furthermore, such techniques primarily focus on group mobility simulation and lack the precision required for reproducing individual paths at the detailed level of GPS data points.
Recent advances endeavor to utilize the generative power of diffusion models to create finer-grained trajectories \cite{zhu2023difftraj}.
Despite their potential, such approaches are unable to control the routing of trajectory generation in accordance to the road network or apply the model directly to a new city without retraining. 
As depicted in \cref{fig:intro}, the current landscape reveals a gap in generating fine-grained trajectories and a lack of control over the outcomes influenced by the road network topology.

Therefore, a desirable trajectory generation solution shall satisfy the following requirements to effectively address the shortcomings of current approaches:
(i) \textbf{High-fidelity}:
The model shall be capable of generating trajectories that are fine-grained and retain the original spatial-temporal properties. 
(ii) \textbf{Flexibility}:
The framework shall be adaptable, enabling users to steer trajectory generation towards simulating specific mobility patterns or complying with distinct conditions, such as constraints imposed by road network topology or varying times of day.
(iii) \textbf{Generalizability}: 
The model shall possess the ability to generalize beyond the geographical context of training data. 
This quality ensures that the generated trajectories remain realistic and applicable when introduced to new environments or used to simulate conditions not directly encountered.

To meet these essential criteria and address the existing gaps, we propose a \textbf{Control}lable \textbf{Traj}ectory (\textbf{\model}) generation framework with the topology-constrained diffusion model.
~\cref{fig:intro} illustrates the superiority of the proposed trajectory generator compared to traditional methods. 
By incorporating road network information, \model can follow its topology structure according to human guidance faithfully, and generate higher resolution trajectories.
In addition, our model showcases remarkable adaptability and generalizability, enabling its application to new cities without the need for retraining.
Firstly, in pursuit of reliable conditional control, we creatively introduce a Masked Road Autoencoder (\textbf{RoadMAE}).
This novel structure allows for capturing fine-grained embedded representations of road segments with real-world scenarios.
Secondly, for the flexibility of condition generation and the transferability across different environments, we develop a geographic attention-based UNet architecture, \textbf{GeoUNet}.
Finally, \model fuses the conditional topology constraints from RoadMAE into GeoUNet to achieve controlled high-fidelity trajectory generation.

In summary, the contributions of our research are as follows:
\begin{itemize}[leftmargin=*]
    \item We design a masked road segment Autoencoder equipped with the ability to encode robust embedding representations based on the given road segment topology.

    \item We introduce a geographic attention-based denoising UNet structure, which seamlessly integrates topological constraints into the diffusion process. This enables the generation of trajectories that are both flexible and controllable.

    \item We validate \model on three real-world trajectory datasets. Quantification and visualization analyses demonstrate that the proposed method can produce high-fidelity trajectories with competitive utility.
    Further, the generated trajectories can be human-controlled and exhibit strong generalizability to unexplored road network topology.
\end{itemize}

\section{Preliminary}\label{sec:pre}


\subsection{Problem Definition}\label{sec:def}

\noindent \textbf{Definition 1} \textbf{(Trajectory)}.
We denote a trajectory as a sequence of continuously sampled GPS points by $\boldsymbol{x}=\{p_i \mid i=1,~2, \ldots, n\}$ with length $n$.
Each GPS point is represented as ${p_i}=\left[\textnormal{lng}_i,~\textnormal{lat}_i,~t_i\right]$, which is a triple of longitude, latitude, and timestamp.

\noindent \textbf{Definition 2} \textbf{(Trip Attributes)}.
Trip attributes are the inherent properties that accompany the entire trajectory. They generally include departure time, travel time, total distance, average speed, etc. 

\noindent \textbf{Definition 3} \textbf{(Road Segments)}.
A road segment is conceptualized as a series of lines formed by two or more connecting GPS points.
Specifically, we denote road segments as $\boldsymbol{r}=\{l_j \mid j=1,2, \ldots, m\}$, where each $l_j = \left[\textnormal{lng}_j^1,~\textnormal{lat}_j^1,~\textnormal{lng}_j^2,~\textnormal{lat}_j^2\right]$ represents a line segment formed by two consecutive GPS points, and $m$ indicates the number of lines. 
For a city, various road segments are interconnected to form the urban road network, i.e., the road topology structure.
In practice, the representation data for such road segments can be obtained from public data sources, e.g., OpenStreetMap \cite{OpenStreetMap}.

\noindent \textbf{Problem Statement} \textbf{(Controllable Trajectory Generation)}.
The primary objective of this research is to develop a controllable trajectory generator $G$ capable of generating multiple trajectories that closely align with given road topology constraints. 
Formally, the problem is defined as follows: Given a set of road segments $\mathcal{R}=\left\{\boldsymbol{r}_1,~\boldsymbol{r}_2, \ldots, \boldsymbol{r}_m\right\}$, the goal for $G$ is to produce a set of generated trajectories $\Tilde{\mathcal{X}}=\left\{\tilde{\boldsymbol{x}}^1,\tilde{\boldsymbol{x}}^2, \ldots, \tilde{\boldsymbol{x}}^n \right\}$, where each trajectory $\tilde{\boldsymbol{x}}^i = \left\{p^{i}_{1},~p^{i}_{2}, \ldots, p^{i}_{n} \right\}$ is a sequence of GPS points. 
The generated trajectories are topology-constrained, realistic, and generalizable.

\begin{figure*}[t]
    \includegraphics[width=0.96\linewidth]{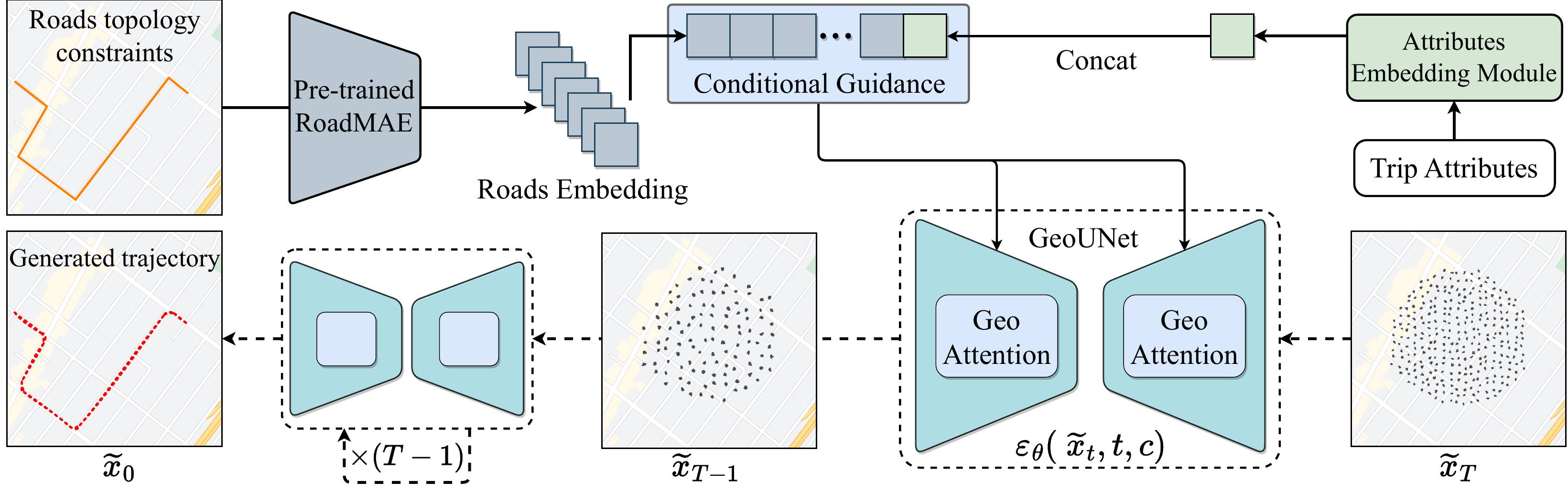}
    \caption{The overview of the proposed \model framework. The pre-trained RoadMAE encodes road embedding based on the topology constraints of the road segments. Road embedding is then concatenated with trip attributes and merged into the diffusion model via geographic attention.}
    \label{fig:overview}
\end{figure*}

\subsection{Conditional Diffusion Probabilistic Model}\label{sec:conddiff}
Diffusion probabilistic models represent a cutting-edge class of generative models renowned for their efficacy in generating complex data distributions \cite{ddpm, ddim, liang2024foundation}. 
These models are particularly effective when conditioned on external information, enabling them to produce data that adheres to specific constraints \cite{wen2023diffstg,hu2023towards}. 
Typically, these models comprise two Markov chain processes with step length $T$:
the forward (noising) process, which incrementally adds noise to the original data $\boldsymbol{x}_0$, and the reverse (denoising) process, which aims to recover the original distribution from a Gaussian noise state $\mathcal{N}(0, \mathbf{I})$.

\textbf{Forward process}.
In the forward process, noise is sequentially added to a dataset of original samples $\boldsymbol{x}_0 \sim q\left(\boldsymbol{x}_0\right)$, where $q\left(\boldsymbol{x}_0\right)$ represents the distribution of the clean data. 
This incremental noising is mathematically defined as a Markov chain that transitions the data from its original state $\boldsymbol{x}_0$ to a noise-dominated state $\boldsymbol{x}_T$:
\begin{align}\label{eq:forward}
    q\left(\boldsymbol{x}_{1: T} \mid \boldsymbol{x}_0\right) &=\prod_{t=1}^T q\left(\boldsymbol{x}_t \mid \boldsymbol{x}_{t-1}\right), \\
        q\left(\boldsymbol{x}_t \mid \boldsymbol{x}_{t-1}\right) &=\mathcal{N}\left(\boldsymbol{x}_t ; \sqrt{1-\beta_t} \boldsymbol{x}_{t-1}, \beta_t \mathbf{I}\right),
\end{align}
where ${\beta_t}$ represents the noise schedule. 
For practical gradient-based optimization, a reparameterization approach \cite{ddpm} is employed such that $\boldsymbol{x}_t = \sqrt{\bar{\alpha}_t} \boldsymbol{x}_0 + \sqrt{1-\bar{\alpha}_t} \boldsymbol{\epsilon}$, and $\bar{\alpha}_t=\prod_{i=1}^{t}(1 - \beta_i)$.
$\boldsymbol{\epsilon} \sim \mathcal{N}(0, \mathbf{I})$ is sampled from a Gaussian distribution: when $T$ is large enough, it ensures that $\boldsymbol{x}_T$ obeys a Gaussian distribution.

\textbf{Reverse process}.
The reverse process of our conditional diffusion probabilistic model is designed to reconstruct the original trajectory data from the noise-corrupted state $\tilde{\boldsymbol{x}}_T$. 
This denoising process depends on the previous noise data and conditional information $\boldsymbol{c}$.
Within the trajectory generation domain, this conditional information encompasses elements such as road segments and trajectory attributes, thereby facilitating the generation of trajectories that are not only noise-free but also conformal to specific control requirements.
Formally, the reverse process is defined by:
\begin{align}\label{eq:reverse_conditional}
p_{\theta}\left(\tilde{\boldsymbol{x}}_{0:T-1} \mid \tilde{\boldsymbol{x}}_{T},  \boldsymbol{c} \right) &= \prod_{t=1}^T p_{\theta}\left(\tilde{\boldsymbol{x}}_{t-1} \mid \tilde{\boldsymbol{x}}_t, \boldsymbol{c}\right), \\
p_{\theta}\left(\tilde{\boldsymbol{x}}_{t-1} \mid \tilde{\boldsymbol{x}}_t, \boldsymbol{c}\right) :&= \mathcal{N}\left(\tilde{\boldsymbol{x}}_{t-1} ; ~\mu_{\theta}\left(\tilde{\boldsymbol{x}}_t, t, \boldsymbol{c} \right), ~\sigma_{\theta}\left(\tilde{\boldsymbol{x}}_t, t, \boldsymbol{c}\right)^2 \mathbf{I}\right),
\end{align}
where $\mu_{\theta}\left(\tilde{\boldsymbol{x}}_t, t, \boldsymbol{c} \right)$ and $\sigma_{\theta}\left(\tilde{\boldsymbol{x}}_t, t, \boldsymbol{c}\right)$ represent the mean and variance of the reverse process at each step $t$, respectively.
Both are parameterized by $\theta$ and modulated by the conditional information $\boldsymbol{c}$.
Building upon the work of Ho \textit{et al.} \cite{ddpm}, we adopt an effective parameterization of $\mu_{\theta}$ and $\sigma_{\theta}$, which are defined as:
\begin{align}\label{eq:contheta}
\left\{\begin{array}{l}
\mu_{\theta}\left(\tilde{\boldsymbol{x}}_t, t, \boldsymbol{c}\right) = \frac{1}{\sqrt{\alpha_t}} \left(\tilde{\boldsymbol{x}}_t - \frac{\beta_t}{\sqrt{1-\bar{\alpha}_t}} \epsilon_{\theta}\left(\tilde{\boldsymbol{x}}_t, t, \boldsymbol{c}\right)\right) \\ \\
\sigma_{\theta}\left(\tilde{\boldsymbol{x}}_t, t, \boldsymbol{c}\right)=\left(\tilde{\beta}_t\right)^{\frac{1}{2}}, \text{where}~~\tilde{\beta}_t = \begin{cases}
\frac{1-\bar{\alpha}_{t-1}}{1-\bar{\alpha}_t} \beta_t &  t > 1 \\
\beta_1 &  t = 1
\end{cases}
\end{array}\right..
\end{align}
Here, $\epsilon_{\theta}\left(\tilde{\boldsymbol{x}}_t, t, \boldsymbol{c}\right)$  denotes the estimated noise level under conditions $t$ and $\boldsymbol{c}$, as predicted by the neural network model uniquely tailored for this task.
Although the diffusion model has been applied in the field of trajectory data generation, prior models did not consider topology constraints of the road network in principle.

\section{Methodology}\label{sec:method}
In this section, we introduce the workflow of our \model framework, designed to produce trajectories that adhere to specific constraints and preferences. \cref{fig:overview} illustrates the entire process, beginning with the generation of road topology constraint guidance through the pre-trained RoadMAE. 
This road embedding, in conjunction with the attribute embedding, informs GeoUNet to ensure the controlled generation of trajectories. GeoUNet, equipped with geo-attention mechanisms, synthesizes the final trajectory by iteratively refining noise estimations conditioned on the embedded road and trip information. 

\subsection{Extracting Topology Constraint }
\begin{figure}[t]
    \includegraphics[width=1\linewidth]{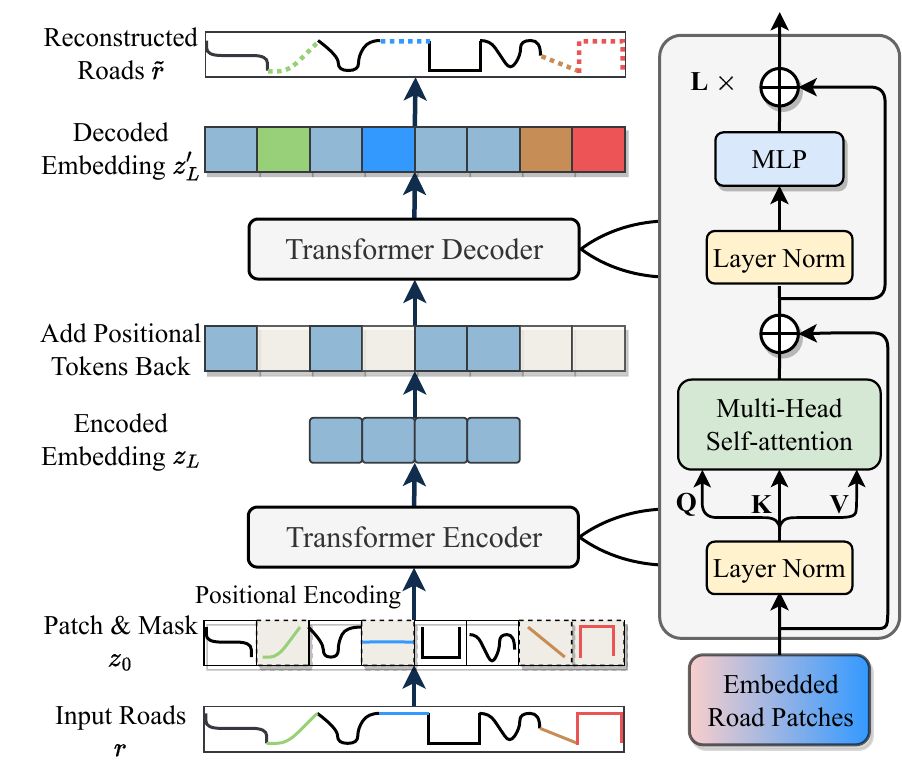}
    \caption{The pipeline of RoadMAE. }
    \label{fig:RoadMAE}
    \vspace*{-6mm}
\end{figure}

When constructing the controllable trajectory generation framework, extracting the conditional guidance is one important prerequisite.
For most diffusion model-based CV or NLP tasks, this is a trivial step easily performed with pre-trained encoders such as Bert \cite{devlin2018bert} or CLIP \cite{radford2021learning}.
However, trajectory generation poses unique challenges due to annotated road network data scarcity and complex geography, such as irregularities and diverse topologies. 
In response, we design a masked road Autoencoder that learns efficiently in a self-supervised manner while capturing essential features of the road network.
~\cref{fig:RoadMAE} showcases the architecture of the proposed RoadMAE, comprising a transformer encoder that processes road patches with positional encoding and a transformer decoder for reconstructing the road segments.
This self-supervised learning framework efficiently distills spatial context into diffusion model-compatible embedding, providing fine-grained guidance for trajectory generation.

\subsubsection{\textbf{Road Segments Patching and Masking}}\label{sec:patchmask}
As defined in \cref{sec:def}, road segments are represented by a series of connected GPS coordinates and maintain geometric and geographic attributes such as shape and location. 
Considering the semantic sparsity at the individual point level, we adopt a patching strategy \cite{ViT} to methodically decompose these segments into discrete, information-rich patches. 
This approach facilitates capturing enough spatial context within patches while mitigating computational demands for lengthy input sequences.
Specifically, we reshape a road segment $\boldsymbol{r} \in \mathbb{R}^{2 \times L}$ into a sequence of patches $\boldsymbol{r}_p \in \mathbb{R}^{N \times (2 \cdot P)}$, where $(2, L)$ represents the original spatial resolution of the road segment, and $P$ denotes the designated patch length.
$N = \lceil L/P \rceil$ signifies the total count of patches. 
Then, we embed each patch into a $D$-dimensional space, and positional embedding are added to retain the sequentiality of the patches. 
The operations above can be formalized as:
\begin{align}
\boldsymbol{z}_0 = \operatorname{Embed}( \operatorname{Patch}(\boldsymbol{r}) )+ \boldsymbol{E}_{pos},
\end{align}
where $\boldsymbol{z}_0 \in \mathbb{R}^{N \times D}$ is the resulting embedded patch sequence and $\boldsymbol{E}_{pos} \in \mathbb{R}^{D}$ is the positional embedding tokens.

In anticipation of potential irregularities and inconsistencies between trajectories and road segments, we introduce a random masking strategy to enhance the robustness of our model, where a subset of patch representations are randomly masked during training \cite{he2022masked}.
This operation forces the encoder to learn a comprehensive understanding of spatial structure, ensuring that it does not overfit specific road segments and enhancing its ability to reconstruct missing information.
Formally, we mathematically represent the masking procedure with a binary mask matrix $\boldsymbol{M} \in \{0,1\}$, where $M_i$ corresponds to the masking state of the $i$-th patch, controlled by a predetermined ratio $r_o$. 
Before fed into transformer, the positioned patch embedding $\boldsymbol{z}_0$ is element-wise multiplied by $\boldsymbol{M}$, producing the masked sequence $\boldsymbol{z}_1 = \boldsymbol{z}_0 \odot \boldsymbol{M}$.
This preparatory step ensures that the encoder can cope with road segment missing situations, facilitating a self-supervised learning paradigm.

\subsubsection{\textbf{Transformer-based Autoencoder}}
Upon preparing the masked sequence of patches, the encoder emerges as a pivotal element in our model, tasked with crafting a contextualized encoding of the road segments. 
The encoder employs a transformer-based architecture (renowned for its efficacy in capturing long-range dependencies) to process sequences through blocks consisting of multi-headed self-attention (MSA), multi-layer perceptron (MLP), and layer normalization (LN) components \cite{vaswani2017attention} (details of the structure can be found in \cref{fig:RoadMAE}).
This architecture empowers the encoder to consider the entire road context when encoding each patch, resulting in a richer, more context-aware representation.
The process can be formalized as:
\begin{align}
\boldsymbol{z}_{\ell}^{\prime}&=\operatorname{MSA}\left(\operatorname{LN}\left(\boldsymbol{z}_{\ell-1}\right)\right)+\boldsymbol{z}_{\ell-1}, \quad \ell=1 \ldots L, \\
\boldsymbol{z}_{\ell}&=\operatorname{MLP}\left(\operatorname{LN}\left(\boldsymbol{z}_{\ell}^{\prime}\right)\right)+\boldsymbol{z}_{\ell}^{\prime},  \quad\quad\quad \ell=1 \ldots L,
\end{align}
where $L$ denotes the number of stacked blocks and output of the encoder $\boldsymbol{z}_L$ is a sequence of coded tensors representing the learned features of the road segments.
It is used by the decoder to reconstruct the road network during the training road segment encoder phase, and as a geo-context condition to control trajectory generation during the trajectory generation phase.

The primary objective of the decoder is to interpret the encoded sequences $\boldsymbol{z}_L$ and reconstruct the segments that were masked during the encoding process. 
It uses the same architecture as the encoder (the number of stacked blocks can be different), which ensures a coherent and symmetric processing flow that facilitates efficient reconstruction of road segments.
Specifically, we first construct a learnable tensor $E_{tmp} \in \mathbb{R}^{1\times D}$, which is merged into the encoded vector based on the position tokens.
Then, the compacted information is effectively decoded back into spatially coherent segments by inverting the transforms, which are expressed as follows:
\begin{align}
\boldsymbol{z}_{L}^{\prime} &= \boldsymbol{z}_{L} \odot \boldsymbol{M} + E_{tmp}  \odot (1 - \boldsymbol{M}), \\
 \tilde{\boldsymbol{r}_i} &= \operatorname{Decoder}(\boldsymbol{z}_{L}^{\prime}).
\end{align}
The efficacy of RoadMAE is optimized through the reconstruction loss $\mathcal{L}_{ssl}$, calculated as the squared error between the original road segments $\boldsymbol{r}_i$ and the reconstructed segments $\tilde{\boldsymbol{r}_i}$:
\begin{align}
\mathcal{L}_{ssl}= \left \| (\boldsymbol{r} -  \tilde{\boldsymbol{r}}) \odot \boldsymbol{M} \right \|^2 .
\end{align}
This function ensures that the Autoencoder focuses on accurately reconstructing the occluded portion of roads, thus enhancing its ability to infer missing information based on learned contextual cues. 
Meanwhile, the encoder is empowered to generate robust geographic representations based on the masked road segments.

\subsection{GeoUNet Architecture}
As discussed in \cref{sec:conddiff}, we need to meticulously design a neural network that can accurately predict the noise level $\epsilon_{\theta}\left(\tilde{\boldsymbol{x}}_t, t, \boldsymbol{c}\right)$ at each reverse diffusion step.
While the UNet architecture has been validated in a range of diffusion models \cite{unet,ddpm}, it still lacks geo-context awareness of trajectory data.
To address this challenge, the proposed GeoUNet integrates geo-attention mechanisms that enable accurate modeling of complex road networks and spatial-temporal dynamics.
Specifically, the designed GeoUNet equips the two primary parts, i.e., downsampling and upsampling, with a devised Geo-Attention mechanism, which is tailored for trajectory processing.
Each part consists of multiple stacked neural network blocks designed to extract and process features at various scales. 
This architecture enables the fusion of multi-scale features through skip connections and allows for efficient capture of both local and global contextual information (More details about GeoUNet can be found in \cref{fig:geounet} of
\cref{app:geounet}).

Then, the geo-attention is integrated into GeoUNet, which augments the model with geo-context embedding at each sampling block. 
In practice, the geo-attention mechanism is implemented by cascading two attention mechanisms:
\begin{align}
    \boldsymbol{Q} &= \boldsymbol{W}_{sq} 
 \cdot \boldsymbol{h}^i, \quad \boldsymbol{K} = \boldsymbol{W}_{sk} 
 \cdot \boldsymbol{h}^i, \quad \boldsymbol{V} = \boldsymbol{W}_{sv} 
 \cdot \boldsymbol{h}^i,\\
     \boldsymbol{Q} &= \boldsymbol{W}_{cq} 
 \cdot \boldsymbol{h}^i, \quad \boldsymbol{K} = \boldsymbol{W}_{ck} 
 \cdot \boldsymbol{c}, ~\quad  \boldsymbol{V} = \boldsymbol{W}_{cv} 
 \cdot \boldsymbol{c}, 
\end{align}
where $\boldsymbol{h} \in \mathbb{R}^{L \times d}$ is the feature embedding of sampling blocks, $\boldsymbol{c}$ is the attribute embedding, and all $\boldsymbol{W}_{s\_} \in \mathbb{R}^{d \times d}$ and $\boldsymbol{W}_{c\_} \in \mathbb{R}^{N \times d}$ are learnable matrices.
Here, the first formula represents the self-attention mechanism, which enables the model to internalize and emphasize fine-grained features in the sequence of trajectories.
Meanwhile, the cross-attention mechanism (second formulation) critically integrates external conditional information such as road segment embedding, trajectory attributes, etc. 
This aligns the features with this additional context to render the final generated trajectories more realistic and context-specific.
Then, the outcomes of both attention mechanisms are computed by $\boldsymbol{h}^{i+1} =  \operatorname{softmax}(\frac{\boldsymbol{Q}  \boldsymbol{K}^T}{\sqrt{d}}) \boldsymbol{V}$.
Finally, the output features are further processed by the standard residual network block in UNet.

In addition, we recognize the critical role of various external attributes (e.g., departure time and travel distance) in shaping the motion patterns and properties of real-world trajectories.
To capture these trajectory properties, we follow the design in literature \cite{cheng2016wide,zhu2023difftraj} by integrating the Attribute embedding module.
In practice, these attribute embedding will be concatenated with the road segment embedding to serve as the conditional information ($\boldsymbol{c} = \operatorname{Concat} (\boldsymbol{z}_{attr}, \boldsymbol{z}_L)$) to guide trajectory generation.

\let\oldnl\nl
\newcommand{\nonl}{\renewcommand{\nl}{\let\nl\oldnl}}
\begin{algorithm}[t]
  \caption{The main processes of \model}\label{alg:diffusion_process}
  \raggedright

\textbf{Training Process:}
\begin{algorithmic} [1]
    \For{ $i = 1, 2, \ldots,$}
      \State \textcolor{purple}{Road segments $\boldsymbol{r}$, Mask ratio $r_o$}, Trajectory Attributes
      \State Get \textcolor{purple}{masked embedding $\boldsymbol{z}_{L}$} by \textcolor{purple}{encoder of RoadMAE}
      \State Get \textcolor{purple}{conditional guidance $\boldsymbol{c} = \operatorname{Concat} (\boldsymbol{z}_{attr}, \boldsymbol{z}_L)$}
      \State Sample $\boldsymbol{x}_0 \sim q(\boldsymbol{x})$,
      \State Sample $t \sim \operatorname{Uniform}(\left\{1, \ldots, T \right\})$, $\epsilon \sim \mathcal{N}(0, \mathbf{I})$
      \State $\boldsymbol{x}_t = \boldsymbol{x}_0 +\sqrt{1-\bar{\alpha}_t} \boldsymbol{\epsilon}$
      \State Updating the gradient $\nabla_{\theta}\left\| \boldsymbol{\epsilon}-\boldsymbol{\epsilon}_\theta\left(\boldsymbol{x}_t,~t,~\boldsymbol{c}\right)\right\|^2_2$
    \EndFor
\end{algorithmic}

\textbf{Generating Process:}
\setcounter{algorithm}{8}
\begin{algorithmic}[1]
    \makeatletter
    \setcounter{ALG@line}{9}
    \State \textcolor{purple}{Road segments $\boldsymbol{r}$}, Trajectory Attributes
    \State Get \textcolor{purple}{road segments embedding $\boldsymbol{z}_{L}$}
    \State Get \textcolor{purple}{conditional guidance $\boldsymbol{c} = \operatorname{Concat} (\boldsymbol{z}_{attr}, \boldsymbol{z}_L)$}
    \State Sample $\tilde{\boldsymbol{x}}_T \sim \mathcal{N}(0,\mathbf{I})$
    \For{ $t = T, T-S, \ldots, 1$}
      \State Compute $\mu_{\theta}\left(\tilde{\boldsymbol{x}}_t, t, \boldsymbol{c}\right)$ according to \cref{eq:contheta}
      \State Compute $p_{\theta}\left(\tilde{\boldsymbol{x}}_{t-1} \mid \tilde{\boldsymbol{x}}_t, \boldsymbol{c}\right)$ according to \cref{eq:reverse_conditional}
    \EndFor \\
    \Return $\tilde{\boldsymbol{x}}_0$
\end{algorithmic}
\end{algorithm}

\subsection{\textbf{Control Trajectory Generation}}
Built on the conditional embedding module and the GeoUNet noise estimation model, the diffusion model allows for controlled trajectory generation based on \cref{eq:contheta}. 
This target is implemented through the training process and the generation process (see \cref{alg:diffusion_process} for details).

\noindent \textbf{Training Process.}
During the training process, the framework first utilizes the encoder of pre-trained RoadMAE and attribute embedding module to convert road segment topology constraints and trip attributes into guideline embedding.
Then, the GeoUNet model estimates the noise level during the reverse process based on the conditional embedding $\boldsymbol{c}$ and the diffusion step $t$.
In this way, the goal of training a diffusion model is to minimize the mean square error between model predicted noise and the Gaussian noise $\epsilon$:
\begin{align}
\min _\theta  \mathcal{L}(\theta)= \min _\theta \mathbb{E}_{\boldsymbol{c}, t, {\boldsymbol{x}}_0 \sim q\left({\boldsymbol{x}}\right), \epsilon \sim \mathcal{N}(0, \mathbf{I}) }\left\| \boldsymbol{\epsilon}-\boldsymbol{\epsilon}_\theta\left(\boldsymbol{x}_t,~t,~\boldsymbol{c}\right)\right\|^2_2.
\end{align}
\noindent \textbf{Generating Process.}
As described in \cref{alg:diffusion_process}, the model is guided by the conditional information to gradually generate trajectories from $\tilde{\boldsymbol{x}}_T \sim \mathcal{N}(0, \mathbf{I})$ in the generating process. 
Within this phase, the desired trajectories are generated from the noise step by step according to \cref{eq:reverse_conditional} and \cref{eq:contheta}.
In addition, we utilize the method proposed in \cite{ddim} to speed up the generating process.
Note that the additional computational overhead of integrating RoadMAE for model training and generation is negligible since we pre-train RoadMAE in advance to obtain topological constraint embedding.
\begin{table*}[t]
\small
    \caption{Performance comparison of \model and generative baselines. }
    \centering
    \begin{tabular}{lccc |ccc |ccc} 
    \toprule
      \multirow{2}{*}[\multirowoffset]{Methods} & \multicolumn{3}{c}{Chengdu} & \multicolumn{3}{c}{Xi'an} & \multicolumn{3}{c}{Porto}  \\
    \cmidrule(lr){2-4}\cmidrule(lr){5-7}\cmidrule(lr){8-10}
     & Density ($\downarrow$) & Trip ($\downarrow$) & Length ($\downarrow$) & Density ($\downarrow$) & Trip ($\downarrow$) & Length ($\downarrow$) & Density ($\downarrow$) & Trip ($\downarrow$) & Length ($\downarrow$) \\ 
    \cmidrule(lr){1-10} 
VAE \cite{Xia2018deeprailway}  & 0.0139 & 0.0502   & 0.0368 & 0.0237 & 0.0608 & 0.0497  & 0.0121 & 0.0224 & 0.0382  \\
TrajGAN \cite{xi2018trajgans} & 0.0137	& 0.0488 & 0.0329 & 0.0220 & 0.0512  & 0.0386 & 0.0101 & 0.0268 & 0.0332 \\
DP-TrajGAN \cite{zhang2022dp} & 0.0127 &	0.0438	& 0.0234  & 0.0207 & 0.0498 & 0.0436 & 0.0109 & 0.0237 & 0.0295 \\
Diffwave \cite{kongdiffwave}  & 0.0145 & 0.0253 & 0.0315 &  0.0213 & 0.0343 & 0.0321 & 0.0106 & 0.0193 & 0.0266  \\
DiffTraj \cite{zhu2023difftraj}  & 0.0066 & 0.0143 & 0.0174 & 0.0126  & 0.0165 & 0.0203 & 0.0087 & 0.0132 & 0.0242 \\
\cmidrule(lr){1-10}
\model w/o $\boldsymbol{c}$ & 0.0079 & 0.0264 & 0.0306  & 0.0141 & 0.0213 & 0.0295 & 0.0114 & 0.0163 & 0.0267 \\
\model-AE  & 0.0047 & 0.0153 & 0.0162 & 0.0127 & 0.0151 & 0.0211 & 0.0071 & 0.0104 & 0.0213 \\
\model  & \textbf{0.0039} & \textbf{0.0106} & \textbf{0.0117} & \textbf{0.0104} & \textbf{0.0125} & \textbf{0.0168} & \textbf{0.0052} & \textbf{0.0096} & \textbf{0.0167} \\
    \bottomrule
    \end{tabular}
    \vspace{0.5em}
    \\\textbf{Bold} indicates the best performance over the baselines. $\downarrow$: lower is better.
    \label{tab:comp}
\end{table*}

\begin{figure*}[h]
    \subfigure[Chengdu]{
    \includegraphics[width=0.485\linewidth]{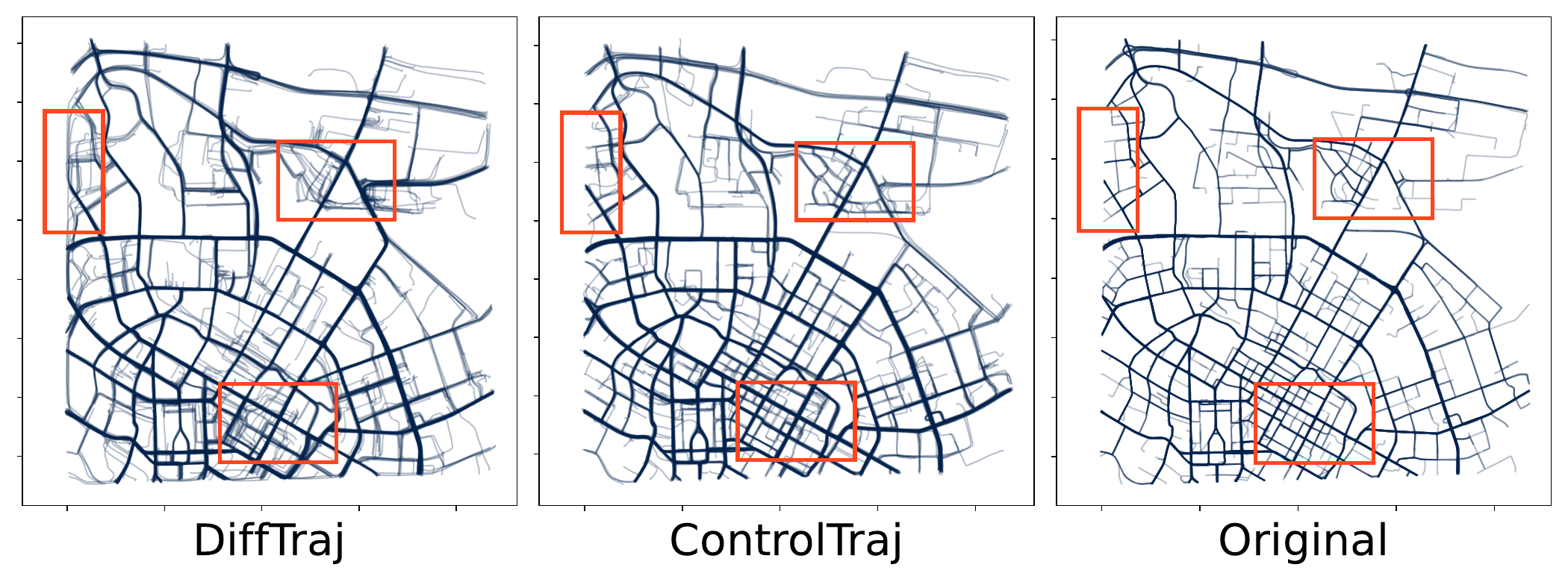}        }
    \subfigure[Xi'an]{
    \includegraphics[width=0.485\linewidth]{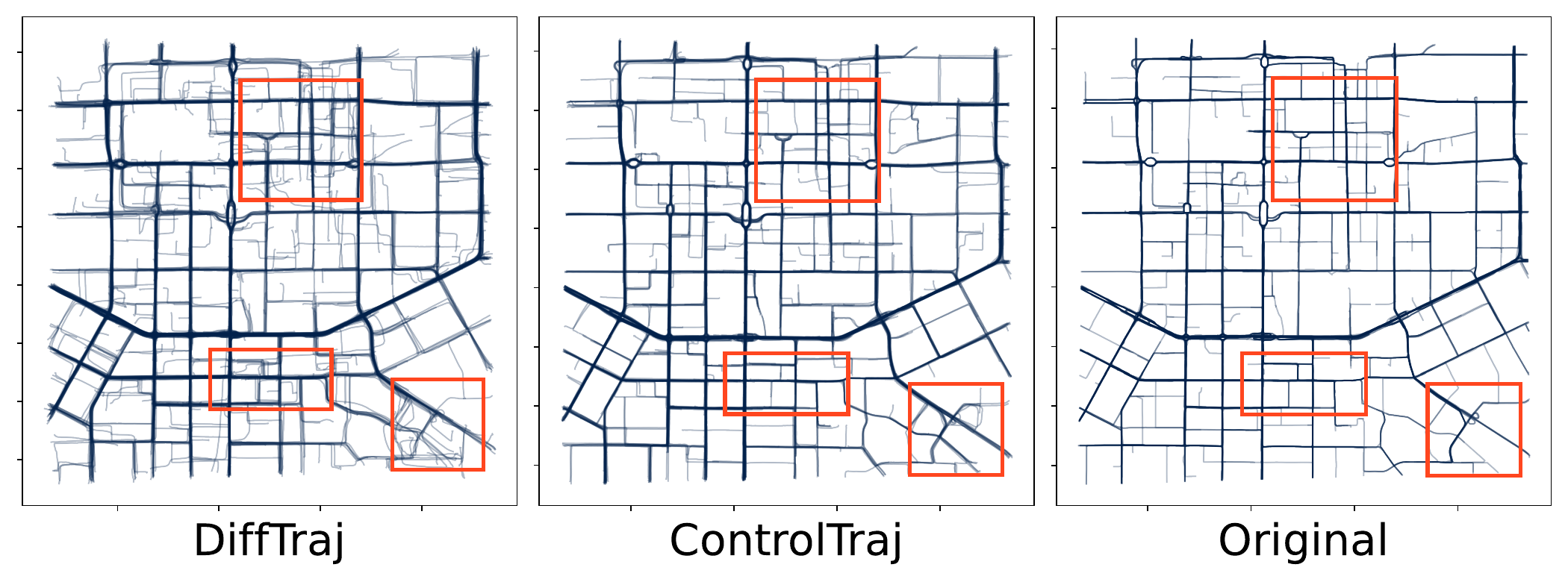}
    }
    \vspace{-3mm}
    \caption{Visualization for different trajectory generation methods. Parameter size: DiffTraj 61.8 MB, ControlTraj 40.7 MB.}
    \label{fig:traj_dis}
\end{figure*}

\section{Experiments}\label{sec:exper}
In this section, we first describe the dataset, baseline, and evaluation metrics setup of the experiment. 
Then, we evaluate the \model framework by comprehensive experiments on real-world datasets to answer the following research questions:

\begin{itemize}[leftmargin=*]
    \item \textbf{RQ1}: Can the trajectory generated by \model provide superior spatial-temporal fidelity compared to several state-of-the-art baselines?
    \item \textbf{RQ2}: How does the realism and utility of the trajectory generated by \model?
    \item \textbf{RQ3}:  Does \model facilitate controllable trajectory generation, allowing for precise manipulation of output trajectories?
    \item \textbf{RQ4}: How does the RoadMAE affect the \model?
    \item \textbf{RQ5}: How do the transferability and generalizability features of \model perform across different geographical contexts?
\end{itemize}

\subsection{Experimental Setups}

\quad \textbf{Datasets.}
We conduct our experiments on three real-world GPS trajectory datasets: Chengdu, Xi'an, and Porto, accessed with appropriate permissions. 
These datasets are rich sources of GPS data, representing diverse urban mobility patterns. 
A comprehensive summary is provided in \textbf{Appendix} \ref{app:dataset} for reference.

\textbf{Baselines.}
To benchmark the performance of our model, we compare it against a suite of leading trajectory generation methods: VAE \cite{Xia2018deeprailway}, TrajGAN \cite{xi2018trajgans}, DP-TrajGAN \cite{zhang2022dp, rao2020lstm}, Diffwave \cite{kongdiffwave}, and DiffTraj \cite{zhu2023difftraj}. 
Moreover, two ablation studies are conducted to evaluate specific components of our model: 1) \model w/o $\boldsymbol{c}$, which examines the contribution of conditional embedding, and 2) \model-AE, where a conventional CNN-based autoencoder substitutes our proposed RoadMAE to assess the impact of different road segment embedding strategies on model efficacy. 
The implementation details are presented in \cref{app:baseline}.

\textbf{Evaluation Metrics.}
We follow the common practice of \cite{du2023ldptrace,zhu2023difftraj} and use three metrics to quantitatively measure the fidelity of the generated trajectories: \textbf{Density error}, \textbf{Travel error}, and \textbf{Length error}.
These metrics collectively evaluate the generated trajectories' geographic distribution, human activity representation, and intra-trajectory coherence, offering insights from both global and local perspectives.
Given that \model can modify trajectory patterns, direct comparison of pattern similarity is deemed inapplicable \cite{zhu2023difftraj}.
Instead, we compute the Jensen-Shannon divergence between distributions of generated and real trajectories, averaging the results over 10 iterations for reliability. 
Further details are available in \cref{app:metrics}.

\textbf{Implement Details.}
The \model implementation involves various general neural network components and parameter settings. 
See \cref{app:imple} for details.

\subsection{Overall Performance (RQ1)}
\quad \textbf{Quantitative analysis.} \cref{tab:comp} summarizes the empirical performance for a suite of baseline methods on diverse real-world datasets.
The results show a clear general trend that distinctly positions \model as a superior solution for trajectory generation across diverse scenarios.
This satisfactory result can be largely attributed to the innovative architecture of \model, which effectively combines geo-attention with road and attribute embedding to enable a fine-grained understanding of spatial-temporal dynamics.
This observation is further validated by a comparison with the most competitive baseline DiffTraj, where the trip error metrics show a substantial improvement.
This remarkable difference suggests that \model, which utilizes detailed road segment embedding, reflects more consistent real-world trajectory behaviors especially for modeling trip dynamics.
Further dissection of \model performance through ablation studies emphasizes the importance of each component.
The performance drop in \model w/o $\boldsymbol{c}$ highlights the critical role of conditional embedding in enhancing the ability to generate contextually accurate trajectories.
Without the road topology constraints, the ability of the model to forge realistic spatial-temporal patterns is diminished.
Similarly, the results for \model-AE are slightly degraded compared to the full \model setup.
This suggests that while the CNN-based autoencoder can capture road segment information, the tailored RoadMAE structure delivers a more robust representation of the road topology.

\textbf{Visualization analysis.} In addition, we also visualize the generation results of DiffTraj and \model to better compare the actual generation capabilities.
\cref{fig:traj_dis} shows the distribution of trajectories generated in Xi'an and Chengdu context (Please refer to \cref{app:geovis} for a larger view and results on other cities).
From the visualization, we can see that both models clearly show a geographic overview of the city.
However, \model presents more fine-grained details.
In particular, \model generates trajectories that are more closely aligned with the underlying road network and conform to the physical constraints within the city for complex or sparse scenarios (marked by red boxes in the figures).
These visual results further confirm the quantitative finding that \model is a superior solution for high-fidelity trajectory generation over existing methods.

\vspace*{-2mm}
\subsection{Generated Trajectories Analysis (RQ2)}
Since the generated trajectories serve to analyze human mobility patterns, their realism and utility are crucial for investigating the data generation method.
In this section, we first present heatmaps of the trajectory distribution within a day for both the generated and original data. Then, we evaluate the utility of the generated data through a well-known traffic flow prediction task.

\begin{figure}[h]
\vspace{-3mm}
    \includegraphics[width=1\linewidth]{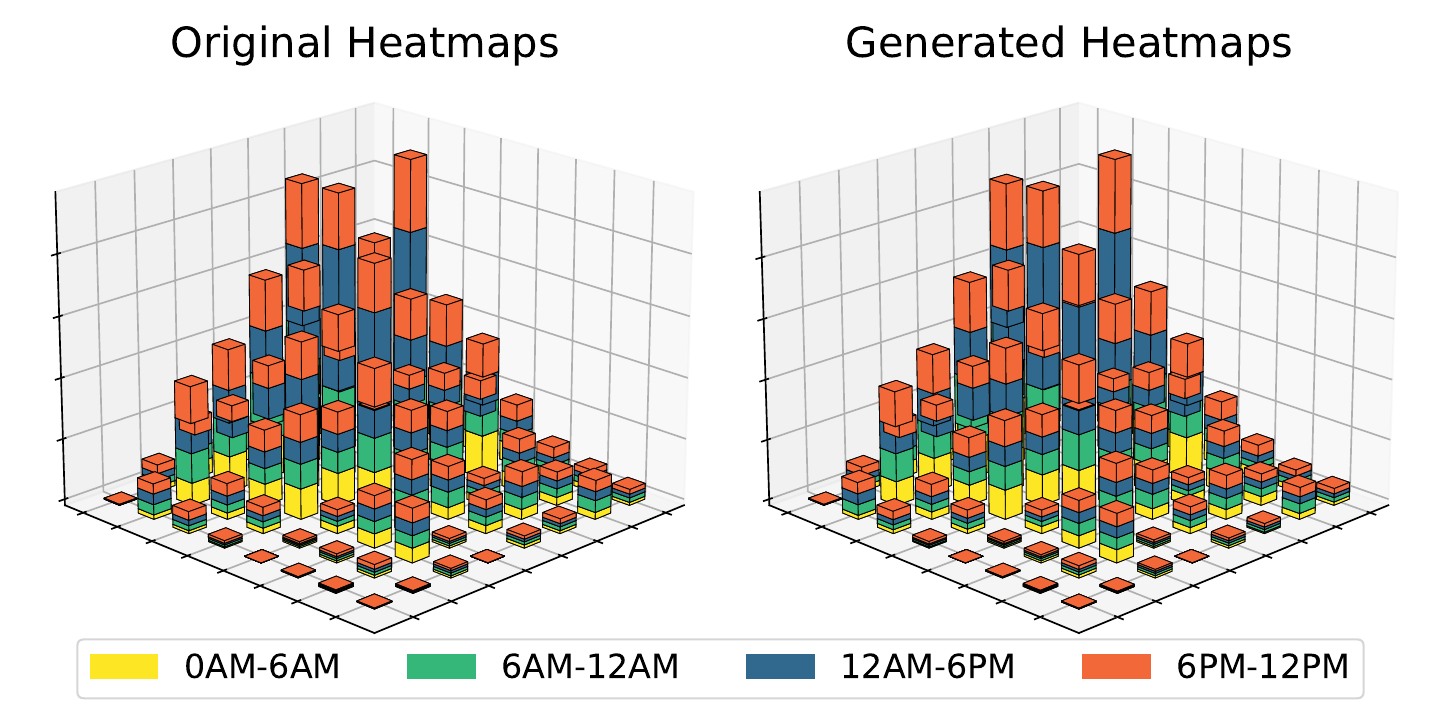}
    \caption{Activity heatmap for periods within a day.}
    \label{fig:heatmap_cd}
\vspace{-3mm}
\end{figure}

The heatmap shown in \cref{fig:heatmap_cd} vividly represents the pattern of human activity at different day periods.
Specifically, colors are used to separate time, bar locations indicate urban areas, and heights reflect activity frequency that provides direct visualization of human activities over the day.
Both heatmaps display prominent activity peaks during expected rush hours, which align with common commuting behaviors. 
For example, the increased activities during the 6 AM--12 PM and 6 PM--12 AM periods reflect the typical morning and evening peaks in urban traffic flow.
Moreover, the consistency of activity levels within different areas further confirms the realism of the generated trajectories.
To summarize, the generated trajectories closely mirror the original distribution in terms of spatial and temporal, suggesting that the generative model has successfully captured the human activity dynamics. 

The utility of the generated data from our \model framework is further substantiated through its downstream application in traffic flow prediction. 
We employed a series of sophisticated neural network models, including ASTGCN \cite{guo2019attention}, GWNet \cite{GWNet}, MTGNN \cite{MTGNN}, and DCRNN \cite{li2017diffusion}, to compare the performance metrics when utilizing the original and the generated data.
Detailed experimental settings and configurations for these tasks are comprehensively documented in \cref{app:analysis}.
The results, summarized in \cref{tab:trafficflow_cd}, are reported as pairs (original / generated) for three evaluation metrics: RMSE, MAE, and MAPE. 
Generally, these metrics reveal a welcomed similarity between the results obtained from the original and the generated data, indicating that the model-generated trajectories retain the spatial-temporal properties required for downstream traffic flow analysis.
For example, the ASTGCN-based models have nearly identical results, and the difference ratios of the performance metrics are usually kept within a moderate range.
These findings emphasize the potential utility of the data generated by \model, evidencing that the generated trajectories exhibit realistic movement patterns that can be exploited in complex predictive models.

\begin{table}[h]
\centering
\small
    \caption{Data utility comparison by traffic flow prediction (Chengdu). Results are expressed as (original / generated).}
\begin{tabular}{lcccc}
\toprule
Methods & ASTGCN & GWNet & MTGNN & DCRNN  \\
\cmidrule{1-5}
RMSE    &  5.76 / 5.75  &  6.76 / 6.48 &  6.82 / 6.40  &  7.29 / 6.76 \\
MAE     &  3.47 / 3.53 &  4.58 / 4.43  & 4.59 / 4.34 &  4.88 / 4.55 \\
MAPE    &  25.47 / 25.58 & 29.39 / 31.70  & 29.73 / 30.79  & 32.40/31.94   \\
\bottomrule
\end{tabular}
\label{tab:trafficflow_cd}
\end{table}

\vspace{-3mm}
\subsection{Controllable Generation (RQ3)}

\begin{figure}[t]
    \subfigure{
    \includegraphics[width=1\linewidth]{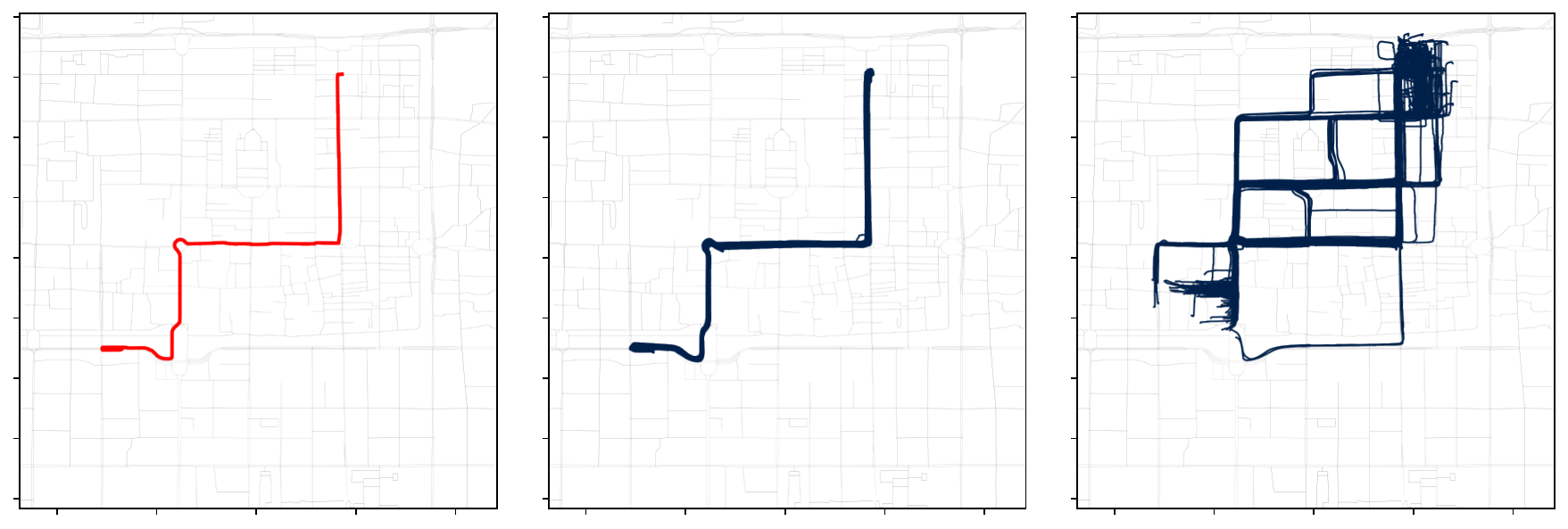}        
    }
    \subfigure{
    \includegraphics[width=1\linewidth]{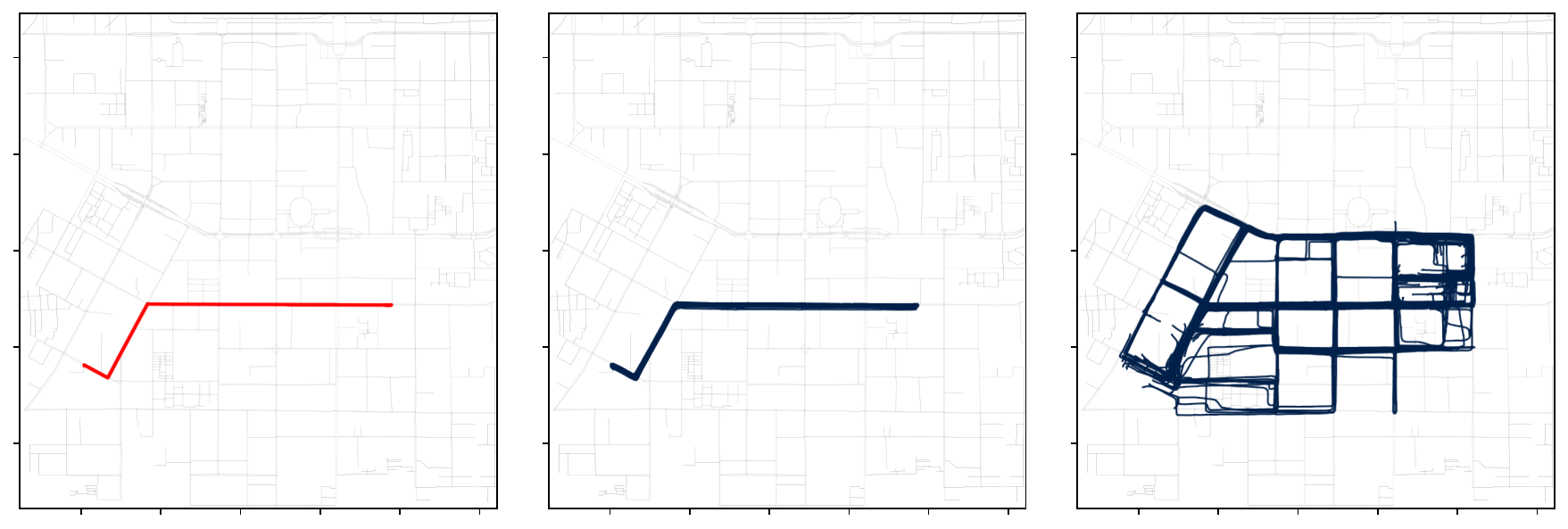}
    }
    \caption{Comparison of controlled trajectory generation. Left: route planning, Middle: \model, Right: DiffTraj.}
    \label{fig:control}
\end{figure}

The capability of controlled trajectory generation is a signature of \model, enabling humans to directly interfere with the traveled geographic route of the trajectory.
Figure \ref{fig:control} illustrates this by comparing the trajectories generated by \model (middle image) and DiffTraj (right image) against the given route planning (left image), with visualizations based on 256 generated trajectories for each model.
The visualization clearly shows that the trajectories generated by the \model adhere closely to predefined routes, showcasing its ability to integrate human-specified instructions.
On the other hand, DiffTraj exhibits limitations in its road guidance capabilities, as it relies solely on the start and end areas for direction \cite{zhu2023difftraj}, leading to a marked variance in the resulting travel routes from the intended path. 
Such divergence from the planned route underscores the shortcomings of DiffTraj in accommodating specific route preferences. 
In contrast, \model not only captures the physical road network accurately and aligns with human-defined routing preferences but also highlights its potential for applications that request custom trajectory planning, from urban planning to navigation systems.
More controlled generation results are presented in \cref{app:control}.

\subsection{Component Analysis (RQ4)}

\begin{figure}[t]
    \subfigure{
    \includegraphics[width=1\linewidth]{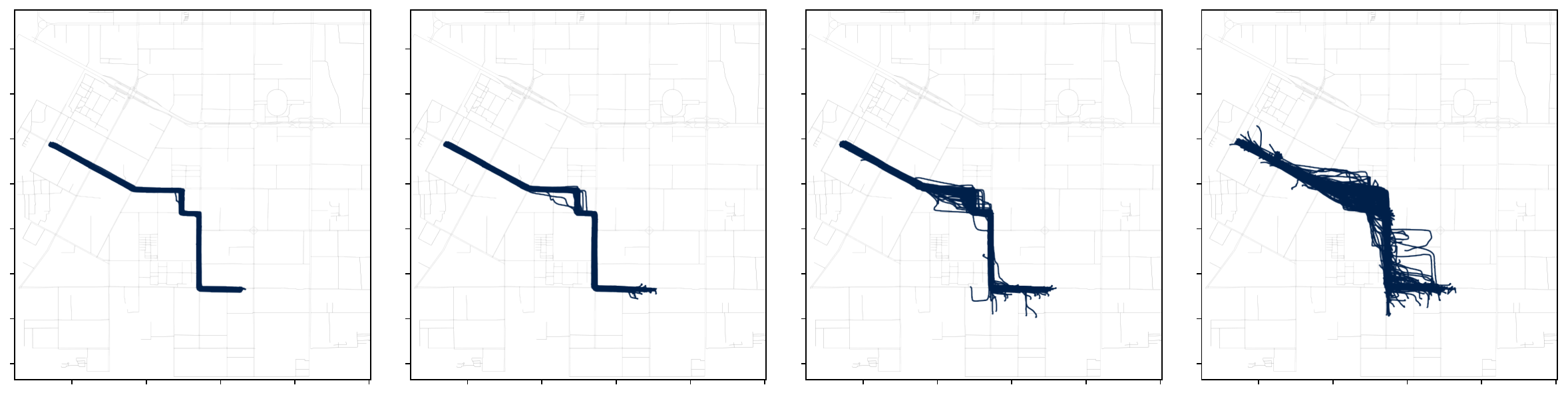}  
    }
    \subfigure{
    \includegraphics[width=1\linewidth]{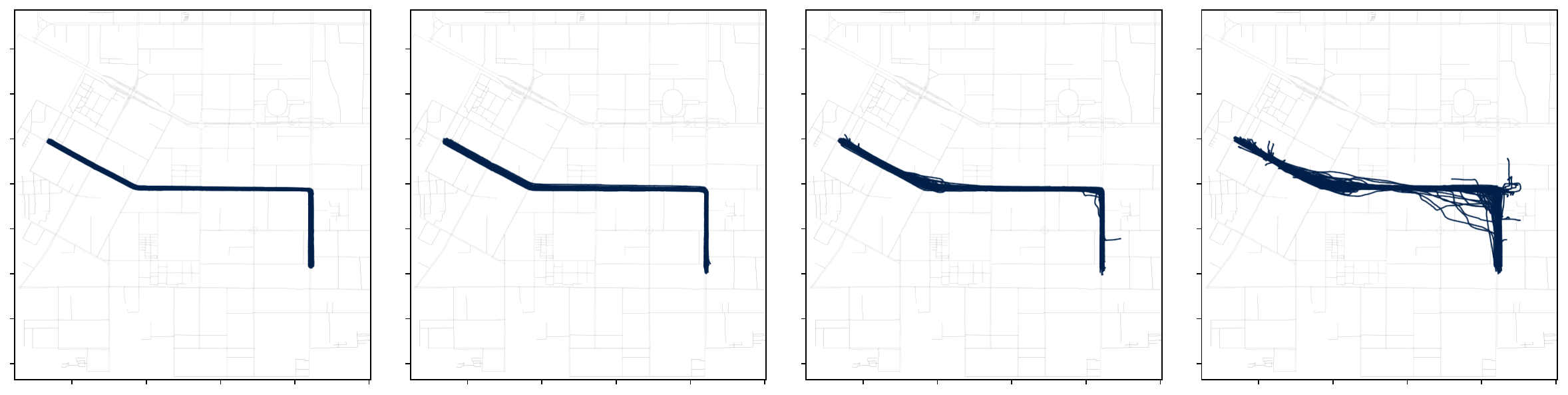}
    }
    \caption{Controllable generation with different mask ratios $\{0, 0.25, 0.5, 0.75 \}$ from left to right.}
    \label{fig:maeanaly}
\end{figure}

As we described in \cref{sec:patchmask}, the proposed RoadMAE integrates a masking strategy to enhance the robustness to cope with the missing part of the road segment.
To evaluate the performance of RoadMAE for fine-grained road embedding, we investigate the influence of mask ratios on the generation capability of ControlTraj.
As illustrated in \cref{fig:maeanaly}, we test varying mask ratio $\{0, 0.25, 0.5, 0.75\}$ and generate 256 trajectories for each scenario, while keeping the rest of the settings the same.
The results indicate that with no masking (0 ratio), the model generates trajectories with high fidelity, closely following the intended path.
As the mask ratio increases to 0.25 and 0.5, RoadMAE demonstrates uncertainty but still produces coherent trajectories that largely respect the road network topology.
However, at an even higher mask ratio at 0.75, the trajectories exhibit deviations from the planned route, weakening the ability of the model to infer the correct road structure.
Nonetheless, the above results still firmly demonstrate the robustness of RoadMAE, which produces satisfactory topological guidance with half of the road segment details missing.
More results are presented in \cref{app:roadmae}

\subsection{Zero-shot Learning Test (RQ5)}
\begin{figure}[h]
    \includegraphics[width=1\linewidth]{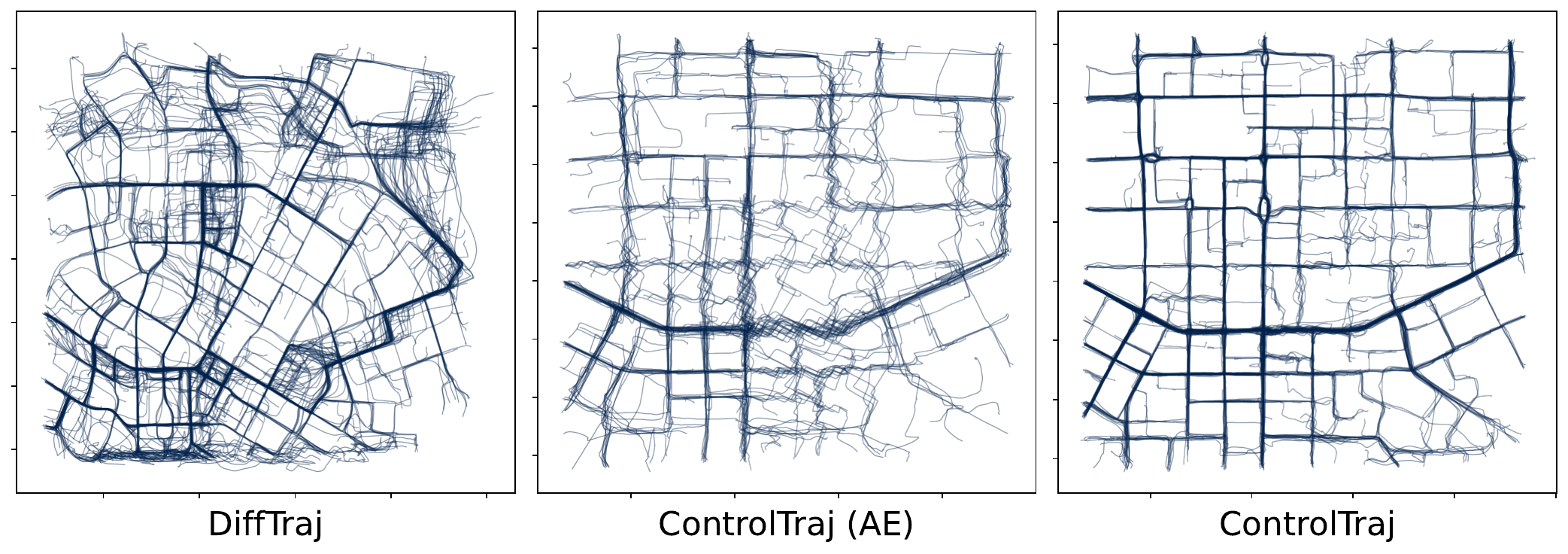}
    \vspace{-5mm}
    \caption{Generalizability of Chengdu $\rightarrow$ Xi'an. (Density error: DiffTraj:	0.0806, ControlTraj (AE): 0.0544, ControlTraj: 0.0171)}
    \label{fig:zero-shot_xa}
\end{figure}

Generalizability and transferability are paramount in evaluating the effectiveness of trajectory generation frameworks, particularly when applied to diverse realistic urban scenarios.
These attributes ensure that models trained with data from one city can seamlessly adapt to new city environments without additional training\,---\,a concept similar to zero-shot learning.
In our experiments, we train the model in one city and then generate trajectories under the guidance of a new city. We compare the performance of three models, DiffTraj, ControlTraj-AE, and ControlTraj, to evaluate their zero-shot learning capabilities.
The results, as visualized in \cref{fig:zero-shot_xa}, reveal that ControlTraj outperforms the other models regarding both transferability and generalizability.
DiffTraj struggles to maintain structural integrity when applied to a new city, as evidenced by its chaotic and unrealistic trajectory patterns. ControlTraj-AE shows improved performance but still exhibits some discrepancies from the road layouts. 
In stark contrast, ControlTraj maintains a high level of accuracy in the generated trajectories, closely reflecting the road network of the new city.
Such performance demonstrates the generalizability and confirms its potential for practical deployment across diverse urban landscapes.
Results on Xi'an $\rightarrow$ Chengdu is shown in \cref{app:generalizability}

\section{Related Work}\label{sec:related}

\quad\textbf{Mobility data generation.}
Broadly speaking, trajectory data synthesis falls into two categories: non-generative and generative~\cite{luca2021survey}. 
Existing research endeavors to generate trajectories that can realistically reflect the spatial-temporal properties and human behavior.
The traditional approach is implemented by statistical models \cite{simini2021deep}, simulations \cite{simini2021deep}, or perturbations \cite{zandbergen2014ensuring}. 
These methods rely on historical data and assumptions about mobility patterns and can provide some level of analysis of human mobility behavior.
With the widespread application of deep neural models such as Generative Adversarial Networks (GANs) \cite{GAN} and Variational Autoencoders (VAEs)\cite{Xia2018deeprailway}, recent research has focused on leveraging them for more complex trajectory modeling.
For example, a number of methods map the trajectory data to a spatial grid and employ a GAN to generate matrixed trajectory data, thus generating trajectories \cite{ouyang2018non, wang2021large,xu2021simulating,feng2020learning,yuan2022activity}.
However, there is a trade-off between its generation accuracy and grid size, only generating coarse-grained behavioral patterns.
Furthermore, some researchers leveraged the image generation ability of GAN by transforming trajectories into images for time-series generation~\cite{cao2021generating, wang2021large}, but the transformation between trajectories and images requires extra computation.
In addition, researchers utilize the image generation capability of GAN to convert trajectories to images for generating ~\cite{cao2021generating, wang2021large}, but the translation between trajectories and images requires additional computation and fails to guarantee fidelity.
A most recent work, DiffTraj~\cite{zhu2023difftraj}, firstly adopted the diffusion model for trajectory synthesizing; however, it lacks controllability and generalizability.

\textbf{Diffusion model.}
As a type of probabilistic generative model, the diffusion model was first introduced by Sohl-Dickstein \textit{et al.}~\cite{sohl2015deep}. It is characterized by two sequential processes, i.e., a forward process that incrementally perturbs the data distribution by adding multi-scale noise, and a reverse process that learns about the data distribution for recovery~\cite{diffusionsuvery}. Numerous attempts have focused on improving the quality of the generative samples and accelerating the sampling speed. DDPM~\cite{ddpm} enhanced it from the aspect of optimization. Song \textit{et al.} proposed a non-Markovian diffusion process to cut down the number of sampling steps~\cite{ddim}, while Nichol \textit{et al.} worked on learning the variance of reverse processes to achieve fewer steps~\cite{nichol2021glide}. Furthermore, Dhariwal \textit{et al.}~\cite{dhariwal2021diffusion} focused on optimizing the structure of the reverse denoising neural network to enhance the sampling quality. Meanwhile, to generate an ideal object, control of the generation process is necessary. The typical method is the conditional diffusion model. For example, DDPM~\cite{ddpm} proposed to add the conditional signal to the attention block in U-Net architecture. Furthermore, Stable Diffusion~\cite{rombach2022high} introduced a cross-attention mechanism to guide the generation based on conditional information, such as texts. 
In this paper, we notably design a masked road segment Autoencoder for real road networks to enable controllable guidance in trajectory generation.

\section{Conclusion}\label{sec:conclusion}
This paper introduces ControlTraj, a novel framework for generating high-fidelity, controllable trajectories through a topology-constrained diffusion model. Our novel approach includes the development of a RoadMAE for capturing detailed embeddings of road segments and a GeoUNet architecture that integrates topological constraints into the diffusion process. Extensively evaluated on three real-world trajectory datasets, ControlTraj has proven its ability to produce trajectories with high fidelity, flexibility, and generalizability.
The results highlight the utility and its potential to simulate realistic human mobility patterns under various constraints.
For future work, we aim to explore the integration of more complex topological dynamics into the ControlTraj framework to further enhance the realism and applicability of the generated trajectories in a wider range of scenarios.

\clearpage
\bibliographystyle{ACM-Reference-Format}
\bibliography{ref}


\begin{thebibliography}{55}


\ifx \showCODEN    \undefined \def \showCODEN     #1{\unskip}     \fi
\ifx \showDOI      \undefined \def \showDOI       #1{#1}\fi
\ifx \showISBNx    \undefined \def \showISBNx     #1{\unskip}     \fi
\ifx \showISBNxiii \undefined \def \showISBNxiii  #1{\unskip}     \fi
\ifx \showISSN     \undefined \def \showISSN      #1{\unskip}     \fi
\ifx \showLCCN     \undefined \def \showLCCN      #1{\unskip}     \fi
\ifx \shownote     \undefined \def \shownote      #1{#1}          \fi
\ifx \showarticletitle \undefined \def \showarticletitle #1{#1}   \fi
\ifx \showURL      \undefined \def \showURL       {\relax}        \fi
\providecommand\bibfield[2]{#2}
\providecommand\bibinfo[2]{#2}
\providecommand\natexlab[1]{#1}
\providecommand\showeprint[2][]{arXiv:#2}

\bibitem[Barbosa et~al\mbox{.}(2018)]%
        {barbosa2018human}
\bibfield{author}{\bibinfo{person}{Hugo Barbosa}, \bibinfo{person}{Marc Barthelemy}, \bibinfo{person}{Gourab Ghoshal}, \bibinfo{person}{Charlotte~R James}, \bibinfo{person}{Maxime Lenormand}, \bibinfo{person}{Thomas Louail}, \bibinfo{person}{Ronaldo Menezes}, \bibinfo{person}{Jos{\'e}~J Ramasco}, \bibinfo{person}{Filippo Simini}, {and} \bibinfo{person}{Marcello Tomasini}.} \bibinfo{year}{2018}\natexlab{}.
\newblock \showarticletitle{Human mobility: Models and applications}.
\newblock \bibinfo{journal}{\emph{Physics Reports}}  \bibinfo{volume}{734} (\bibinfo{year}{2018}), \bibinfo{pages}{1--74}.
\newblock


\bibitem[Cao and Li(2021)]%
        {cao2021generating}
\bibfield{author}{\bibinfo{person}{Chu Cao} {and} \bibinfo{person}{Mo Li}.} \bibinfo{year}{2021}\natexlab{}.
\newblock \showarticletitle{Generating Mobility Trajectories with Retained Data Utility}. In \bibinfo{booktitle}{\emph{Proceedings of the 27th ACM SIGKDD Conference on Knowledge Discovery \& Data Mining}}. \bibinfo{pages}{2610--2620}.
\newblock


\bibitem[Chen et~al\mbox{.}(2024)]%
        {chen2024deep}
\bibfield{author}{\bibinfo{person}{Wei Chen}, \bibinfo{person}{Yuxuan Liang}, \bibinfo{person}{Yuanshao Zhu}, \bibinfo{person}{Yanchuan Chang}, \bibinfo{person}{Kang Luo}, \bibinfo{person}{Haomin Wen}, \bibinfo{person}{Lei Li}, \bibinfo{person}{Yanwei Yu}, \bibinfo{person}{Qingsong Wen}, \bibinfo{person}{Chao Chen}, {et~al\mbox{.}}} \bibinfo{year}{2024}\natexlab{}.
\newblock \showarticletitle{Deep Learning for Trajectory Data Management and Mining: A Survey and Beyond}.
\newblock \bibinfo{journal}{\emph{arXiv preprint arXiv:2403.14151}} (\bibinfo{year}{2024}).
\newblock


\bibitem[Cheng et~al\mbox{.}(2016)]%
        {cheng2016wide}
\bibfield{author}{\bibinfo{person}{Heng-Tze Cheng}, \bibinfo{person}{Levent Koc}, \bibinfo{person}{Jeremiah Harmsen}, \bibinfo{person}{Tal Shaked}, \bibinfo{person}{Tushar Chandra}, \bibinfo{person}{Hrishi Aradhye}, \bibinfo{person}{Glen Anderson}, \bibinfo{person}{Greg Corrado}, \bibinfo{person}{Wei Chai}, \bibinfo{person}{Mustafa Ispir}, {et~al\mbox{.}}} \bibinfo{year}{2016}\natexlab{}.
\newblock \showarticletitle{Wide \& deep learning for recommender systems}. In \bibinfo{booktitle}{\emph{Proceedings of the 1st workshop on deep learning for recommender systems}}. \bibinfo{pages}{7--10}.
\newblock


\bibitem[Dai et~al\mbox{.}(2015)]%
        {dai2015personalized}
\bibfield{author}{\bibinfo{person}{Jian Dai}, \bibinfo{person}{Bin Yang}, \bibinfo{person}{Chenjuan Guo}, {and} \bibinfo{person}{Zhiming Ding}.} \bibinfo{year}{2015}\natexlab{}.
\newblock \showarticletitle{Personalized route recommendation using big trajectory data}. In \bibinfo{booktitle}{\emph{2015 IEEE 31st international conference on data engineering}}. IEEE, \bibinfo{pages}{543--554}.
\newblock


\bibitem[Devlin et~al\mbox{.}(2018)]%
        {devlin2018bert}
\bibfield{author}{\bibinfo{person}{Jacob Devlin}, \bibinfo{person}{Ming-Wei Chang}, \bibinfo{person}{Kenton Lee}, {and} \bibinfo{person}{Kristina Toutanova}.} \bibinfo{year}{2018}\natexlab{}.
\newblock \showarticletitle{Bert: Pre-training of deep bidirectional transformers for language understanding}.
\newblock \bibinfo{journal}{\emph{arXiv preprint arXiv:1810.04805}} (\bibinfo{year}{2018}).
\newblock


\bibitem[Dhariwal and Nichol(2021)]%
        {dhariwal2021diffusion}
\bibfield{author}{\bibinfo{person}{Prafulla Dhariwal} {and} \bibinfo{person}{Alexander Nichol}.} \bibinfo{year}{2021}\natexlab{}.
\newblock \showarticletitle{Diffusion models beat gans on image synthesis}.
\newblock \bibinfo{journal}{\emph{Advances in Neural Information Processing Systems}}  \bibinfo{volume}{34} (\bibinfo{year}{2021}), \bibinfo{pages}{8780--8794}.
\newblock


\bibitem[Dosovitskiy et~al\mbox{.}(2020)]%
        {ViT}
\bibfield{author}{\bibinfo{person}{Alexey Dosovitskiy}, \bibinfo{person}{Lucas Beyer}, \bibinfo{person}{Alexander Kolesnikov}, \bibinfo{person}{Dirk Weissenborn}, \bibinfo{person}{Xiaohua Zhai}, \bibinfo{person}{Thomas Unterthiner}, \bibinfo{person}{Mostafa Dehghani}, \bibinfo{person}{Matthias Minderer}, \bibinfo{person}{Georg Heigold}, \bibinfo{person}{Sylvain Gelly}, {et~al\mbox{.}}} \bibinfo{year}{2020}\natexlab{}.
\newblock \showarticletitle{An image is worth 16x16 words: Transformers for image recognition at scale}.
\newblock \bibinfo{journal}{\emph{arXiv preprint arXiv:2010.11929}} (\bibinfo{year}{2020}).
\newblock


\bibitem[Du et~al\mbox{.}(2023)]%
        {du2023ldptrace}
\bibfield{author}{\bibinfo{person}{Yuntao Du}, \bibinfo{person}{Yujia Hu}, \bibinfo{person}{Zhikun Zhang}, \bibinfo{person}{Ziquan Fang}, \bibinfo{person}{Lu Chen}, \bibinfo{person}{Baihua Zheng}, {and} \bibinfo{person}{Yunjun Gao}.} \bibinfo{year}{2023}\natexlab{}.
\newblock \showarticletitle{LDPTrace: Locally Differentially Private Trajectory Synthesis}.
\newblock \bibinfo{journal}{\emph{arXiv preprint arXiv:2302.06180}} (\bibinfo{year}{2023}).
\newblock


\bibitem[Fang et~al\mbox{.}(2021)]%
        {fang2021dragoon}
\bibfield{author}{\bibinfo{person}{Ziquan Fang}, \bibinfo{person}{Lu Chen}, \bibinfo{person}{Yunjun Gao}, \bibinfo{person}{Lu Pan}, {and} \bibinfo{person}{Christian~S Jensen}.} \bibinfo{year}{2021}\natexlab{}.
\newblock \showarticletitle{Dragoon: a hybrid and efficient big trajectory management system for offline and online analytics}.
\newblock \bibinfo{journal}{\emph{The VLDB Journal}}  \bibinfo{volume}{30} (\bibinfo{year}{2021}), \bibinfo{pages}{287--310}.
\newblock


\bibitem[Feng et~al\mbox{.}(2020)]%
        {feng2020learning}
\bibfield{author}{\bibinfo{person}{Jie Feng}, \bibinfo{person}{Zeyu Yang}, \bibinfo{person}{Fengli Xu}, \bibinfo{person}{Haisu Yu}, \bibinfo{person}{Mudan Wang}, {and} \bibinfo{person}{Yong Li}.} \bibinfo{year}{2020}\natexlab{}.
\newblock \showarticletitle{Learning to simulate human mobility}. In \bibinfo{booktitle}{\emph{Proceedings of the 26th ACM SIGKDD international conference on knowledge discovery \& data mining}}. \bibinfo{pages}{3426--3433}.
\newblock


\bibitem[Goodfellow et~al\mbox{.}(2014)]%
        {GAN}
\bibfield{author}{\bibinfo{person}{Ian Goodfellow}, \bibinfo{person}{Jean Pouget-Abadie}, \bibinfo{person}{Mehdi Mirza}, \bibinfo{person}{Bing Xu}, \bibinfo{person}{David Warde-Farley}, \bibinfo{person}{Sherjil Ozair}, \bibinfo{person}{Aaron Courville}, {and} \bibinfo{person}{Yoshua Bengio}.} \bibinfo{year}{2014}\natexlab{}.
\newblock \showarticletitle{Generative Adversarial Nets}. In \bibinfo{booktitle}{\emph{Advances in Neural Information Processing Systems}}, Vol.~\bibinfo{volume}{27}.
\newblock


\bibitem[Guo et~al\mbox{.}(2018)]%
        {guo2018learning}
\bibfield{author}{\bibinfo{person}{Chenjuan Guo}, \bibinfo{person}{Bin Yang}, \bibinfo{person}{Jilin Hu}, {and} \bibinfo{person}{Christian Jensen}.} \bibinfo{year}{2018}\natexlab{}.
\newblock \showarticletitle{Learning to route with sparse trajectory sets}. In \bibinfo{booktitle}{\emph{2018 IEEE 34th International Conference on Data Engineering (ICDE)}}. IEEE, \bibinfo{pages}{1073--1084}.
\newblock


\bibitem[Guo et~al\mbox{.}(2019)]%
        {guo2019attention}
\bibfield{author}{\bibinfo{person}{Shengnan Guo}, \bibinfo{person}{Youfang Lin}, \bibinfo{person}{Ning Feng}, \bibinfo{person}{Chao Song}, {and} \bibinfo{person}{Huaiyu Wan}.} \bibinfo{year}{2019}\natexlab{}.
\newblock \showarticletitle{Attention based spatial-temporal graph convolutional networks for traffic flow forecasting}. In \bibinfo{booktitle}{\emph{Proceedings of the AAAI conference on artificial intelligence}}, Vol.~\bibinfo{volume}{33}. \bibinfo{pages}{922--929}.
\newblock


\bibitem[He et~al\mbox{.}(2022)]%
        {he2022masked}
\bibfield{author}{\bibinfo{person}{Kaiming He}, \bibinfo{person}{Xinlei Chen}, \bibinfo{person}{Saining Xie}, \bibinfo{person}{Yanghao Li}, \bibinfo{person}{Piotr Doll{\'a}r}, {and} \bibinfo{person}{Ross Girshick}.} \bibinfo{year}{2022}\natexlab{}.
\newblock \showarticletitle{Masked autoencoders are scalable vision learners}. In \bibinfo{booktitle}{\emph{Proceedings of the IEEE/CVF conference on computer vision and pattern recognition}}. \bibinfo{pages}{16000--16009}.
\newblock


\bibitem[Henke et~al\mbox{.}(2023)]%
        {henke2023condtraj}
\bibfield{author}{\bibinfo{person}{Nils Henke}, \bibinfo{person}{Shimon Wonsak}, \bibinfo{person}{Prasenjit Mitra}, \bibinfo{person}{Michael Nolting}, {and} \bibinfo{person}{Nicolas Tempelmeier}.} \bibinfo{year}{2023}\natexlab{}.
\newblock \showarticletitle{Condtraj-gan: Conditional sequential gan for generating synthetic vehicle trajectories}. In \bibinfo{booktitle}{\emph{Pacific-Asia Conference on Knowledge Discovery and Data Mining}}. Springer, \bibinfo{pages}{79--91}.
\newblock


\bibitem[Ho et~al\mbox{.}(2020)]%
        {ddpm}
\bibfield{author}{\bibinfo{person}{Jonathan Ho}, \bibinfo{person}{Ajay Jain}, {and} \bibinfo{person}{Pieter Abbeel}.} \bibinfo{year}{2020}\natexlab{}.
\newblock \showarticletitle{Denoising diffusion probabilistic models}.
\newblock \bibinfo{journal}{\emph{Advances in Neural Information Processing Systems}}  \bibinfo{volume}{33} (\bibinfo{year}{2020}), \bibinfo{pages}{6840--6851}.
\newblock


\bibitem[Hu et~al\mbox{.}(2023)]%
        {hu2023towards}
\bibfield{author}{\bibinfo{person}{Junfeng Hu}, \bibinfo{person}{Xu Liu}, \bibinfo{person}{Zhencheng Fan}, \bibinfo{person}{Yuxuan Liang}, {and} \bibinfo{person}{Roger Zimmermann}.} \bibinfo{year}{2023}\natexlab{}.
\newblock \showarticletitle{Towards Unifying Diffusion Models for Probabilistic Spatio-Temporal Graph Learning}.
\newblock \bibinfo{journal}{\emph{arXiv preprint arXiv:2310.17360}} (\bibinfo{year}{2023}).
\newblock


\bibitem[Jensen(2022)]%
        {jensen2022digitalization}
\bibfield{author}{\bibinfo{person}{Christian~S Jensen}.} \bibinfo{year}{2022}\natexlab{}.
\newblock \showarticletitle{Digitalization in the Service of Society: The Case of Big Vehicle Trajectory Data}. In \bibinfo{booktitle}{\emph{Proceedings of the 34th International Conference on Scientific and Statistical Database Management}}. \bibinfo{pages}{1--1}.
\newblock


\bibitem[Kong et~al\mbox{.}(2020)]%
        {kongdiffwave}
\bibfield{author}{\bibinfo{person}{Zhifeng Kong}, \bibinfo{person}{Wei Ping}, \bibinfo{person}{Jiaji Huang}, \bibinfo{person}{Kexin Zhao}, {and} \bibinfo{person}{Bryan Catanzaro}.} \bibinfo{year}{2020}\natexlab{}.
\newblock \showarticletitle{DiffWave: A Versatile Diffusion Model for Audio Synthesis}. In \bibinfo{booktitle}{\emph{International Conference on Learning Representations}}.
\newblock


\bibitem[Li et~al\mbox{.}(2017)]%
        {li2017diffusion}
\bibfield{author}{\bibinfo{person}{Yaguang Li}, \bibinfo{person}{Rose Yu}, \bibinfo{person}{Cyrus Shahabi}, {and} \bibinfo{person}{Yan Liu}.} \bibinfo{year}{2017}\natexlab{}.
\newblock \showarticletitle{Diffusion convolutional recurrent neural network: Data-driven traffic forecasting}.
\newblock \bibinfo{journal}{\emph{arXiv preprint arXiv:1707.01926}} (\bibinfo{year}{2017}).
\newblock


\bibitem[Liang et~al\mbox{.}(2021)]%
        {trajode}
\bibfield{author}{\bibinfo{person}{Yuxuan Liang}, \bibinfo{person}{Kun Ouyang}, \bibinfo{person}{Hanshu Yan}, \bibinfo{person}{Yiwei Wang}, \bibinfo{person}{Zekun Tong}, {and} \bibinfo{person}{Roger Zimmermann}.} \bibinfo{year}{2021}\natexlab{}.
\newblock \showarticletitle{Modeling Trajectories with Neural Ordinary Differential Equations}. In \bibinfo{booktitle}{\emph{IJCAI}}. \bibinfo{pages}{1498--1504}.
\newblock


\bibitem[Liang et~al\mbox{.}(2024)]%
        {liang2024foundation}
\bibfield{author}{\bibinfo{person}{Yuxuan Liang}, \bibinfo{person}{Haomin Wen}, \bibinfo{person}{Yuqi Nie}, \bibinfo{person}{Yushan Jiang}, \bibinfo{person}{Ming Jin}, \bibinfo{person}{Dongjin Song}, \bibinfo{person}{Shirui Pan}, {and} \bibinfo{person}{Qingsong Wen}.} \bibinfo{year}{2024}\natexlab{}.
\newblock \showarticletitle{Foundation Models for Time Series Analysis: A Tutorial and Survey}.
\newblock \bibinfo{journal}{\emph{arXiv preprint arXiv:2403.14735}} (\bibinfo{year}{2024}).
\newblock


\bibitem[Luca et~al\mbox{.}(2021)]%
        {luca2021survey}
\bibfield{author}{\bibinfo{person}{Massimiliano Luca}, \bibinfo{person}{Gianni Barlacchi}, \bibinfo{person}{Bruno Lepri}, {and} \bibinfo{person}{Luca Pappalardo}.} \bibinfo{year}{2021}\natexlab{}.
\newblock \showarticletitle{A survey on deep learning for human mobility}.
\newblock \bibinfo{journal}{\emph{ACM Computing Surveys (CSUR)}} \bibinfo{volume}{55}, \bibinfo{number}{1} (\bibinfo{year}{2021}), \bibinfo{pages}{1--44}.
\newblock


\bibitem[Nichol et~al\mbox{.}(2021)]%
        {nichol2021glide}
\bibfield{author}{\bibinfo{person}{Alex Nichol}, \bibinfo{person}{Prafulla Dhariwal}, \bibinfo{person}{Aditya Ramesh}, \bibinfo{person}{Pranav Shyam}, \bibinfo{person}{Pamela Mishkin}, \bibinfo{person}{Bob McGrew}, \bibinfo{person}{Ilya Sutskever}, {and} \bibinfo{person}{Mark Chen}.} \bibinfo{year}{2021}\natexlab{}.
\newblock \showarticletitle{Glide: Towards photorealistic image generation and editing with text-guided diffusion models}.
\newblock \bibinfo{journal}{\emph{arXiv preprint arXiv:2112.10741}} (\bibinfo{year}{2021}).
\newblock


\bibitem[{OpenStreetMap contributors}(2017)]%
        {OpenStreetMap}
\bibfield{author}{\bibinfo{person}{{OpenStreetMap contributors}}.} \bibinfo{year}{2017}\natexlab{}.
\newblock \bibinfo{title}{{Planet dump retrieved from https://planet.osm.org }}.
\newblock \bibinfo{howpublished}{\url{ https://www.openstreetmap.org }}.
\newblock


\bibitem[Ouyang et~al\mbox{.}(2018)]%
        {ouyang2018non}
\bibfield{author}{\bibinfo{person}{Kun Ouyang}, \bibinfo{person}{Reza Shokri}, \bibinfo{person}{David~S Rosenblum}, {and} \bibinfo{person}{Wenzhuo Yang}.} \bibinfo{year}{2018}\natexlab{}.
\newblock \showarticletitle{A non-parametric generative model for human trajectories}. In \bibinfo{booktitle}{\emph{IJCAI}}, Vol.~\bibinfo{volume}{18}. \bibinfo{pages}{3812--3817}.
\newblock


\bibitem[Radford et~al\mbox{.}(2021)]%
        {radford2021learning}
\bibfield{author}{\bibinfo{person}{Alec Radford}, \bibinfo{person}{Jong~Wook Kim}, \bibinfo{person}{Chris Hallacy}, \bibinfo{person}{Aditya Ramesh}, \bibinfo{person}{Gabriel Goh}, \bibinfo{person}{Sandhini Agarwal}, \bibinfo{person}{Girish Sastry}, \bibinfo{person}{Amanda Askell}, \bibinfo{person}{Pamela Mishkin}, \bibinfo{person}{Jack Clark}, {et~al\mbox{.}}} \bibinfo{year}{2021}\natexlab{}.
\newblock \showarticletitle{Learning transferable visual models from natural language supervision}. In \bibinfo{booktitle}{\emph{International conference on machine learning}}. PMLR, \bibinfo{pages}{8748--8763}.
\newblock


\bibitem[Rao et~al\mbox{.}(2020)]%
        {rao2020lstm}
\bibfield{author}{\bibinfo{person}{Jinmeng Rao}, \bibinfo{person}{Song Gao}, \bibinfo{person}{Yuhao Kang}, {and} \bibinfo{person}{Qunying Huang}.} \bibinfo{year}{2020}\natexlab{}.
\newblock \showarticletitle{{LSTM-TrajGAN}: A Deep Learning Approach to Trajectory Privacy Protection}. In \bibinfo{booktitle}{\emph{11th International Conference on Geographic Information Science (GIScience 2021)}}.
\newblock


\bibitem[Rombach et~al\mbox{.}(2022)]%
        {rombach2022high}
\bibfield{author}{\bibinfo{person}{Robin Rombach}, \bibinfo{person}{Andreas Blattmann}, \bibinfo{person}{Dominik Lorenz}, \bibinfo{person}{Patrick Esser}, {and} \bibinfo{person}{Bj{\"o}rn Ommer}.} \bibinfo{year}{2022}\natexlab{}.
\newblock \showarticletitle{High-resolution image synthesis with latent diffusion models}. In \bibinfo{booktitle}{\emph{Proceedings of the IEEE/CVF Conference on Computer Vision and Pattern Recognition}}. \bibinfo{pages}{10684--10695}.
\newblock


\bibitem[Ronneberger et~al\mbox{.}(2015)]%
        {unet}
\bibfield{author}{\bibinfo{person}{Olaf Ronneberger}, \bibinfo{person}{Philipp Fischer}, {and} \bibinfo{person}{Thomas Brox}.} \bibinfo{year}{2015}\natexlab{}.
\newblock \showarticletitle{U-net: Convolutional networks for biomedical image segmentation}. In \bibinfo{booktitle}{\emph{International Conference on Medical image computing and computer-assisted intervention}}. Springer, \bibinfo{pages}{234--241}.
\newblock


\bibitem[Ruan et~al\mbox{.}(2020)]%
        {ruan2020learning}
\bibfield{author}{\bibinfo{person}{Sijie Ruan}, \bibinfo{person}{Cheng Long}, \bibinfo{person}{Jie Bao}, \bibinfo{person}{Chunyang Li}, \bibinfo{person}{Zisheng Yu}, \bibinfo{person}{Ruiyuan Li}, \bibinfo{person}{Yuxuan Liang}, \bibinfo{person}{Tianfu He}, {and} \bibinfo{person}{Yu Zheng}.} \bibinfo{year}{2020}\natexlab{}.
\newblock \showarticletitle{Learning to generate maps from trajectories}. In \bibinfo{booktitle}{\emph{Proceedings of the AAAI conference on artificial intelligence}}, Vol.~\bibinfo{volume}{34}. \bibinfo{pages}{890--897}.
\newblock


\bibitem[Simini et~al\mbox{.}(2021)]%
        {simini2021deep}
\bibfield{author}{\bibinfo{person}{Filippo Simini}, \bibinfo{person}{Gianni Barlacchi}, \bibinfo{person}{Massimilano Luca}, {and} \bibinfo{person}{Luca Pappalardo}.} \bibinfo{year}{2021}\natexlab{}.
\newblock \showarticletitle{A deep gravity model for mobility flows generation}.
\newblock \bibinfo{journal}{\emph{Nature communications}} \bibinfo{volume}{12}, \bibinfo{number}{1} (\bibinfo{year}{2021}), \bibinfo{pages}{6576}.
\newblock


\bibitem[Sohl-Dickstein et~al\mbox{.}(2015)]%
        {sohl2015deep}
\bibfield{author}{\bibinfo{person}{Jascha Sohl-Dickstein}, \bibinfo{person}{Eric Weiss}, \bibinfo{person}{Niru Maheswaranathan}, {and} \bibinfo{person}{Surya Ganguli}.} \bibinfo{year}{2015}\natexlab{}.
\newblock \showarticletitle{Deep unsupervised learning using nonequilibrium thermodynamics}. In \bibinfo{booktitle}{\emph{International Conference on Machine Learning}}. PMLR, \bibinfo{pages}{2256--2265}.
\newblock


\bibitem[Song et~al\mbox{.}(2020)]%
        {ddim}
\bibfield{author}{\bibinfo{person}{Jiaming Song}, \bibinfo{person}{Chenlin Meng}, {and} \bibinfo{person}{Stefano Ermon}.} \bibinfo{year}{2020}\natexlab{}.
\newblock \showarticletitle{Denoising diffusion implicit models}.
\newblock \bibinfo{journal}{\emph{arXiv preprint arXiv:2010.02502}} (\bibinfo{year}{2020}).
\newblock


\bibitem[Vaswani et~al\mbox{.}(2017)]%
        {vaswani2017attention}
\bibfield{author}{\bibinfo{person}{Ashish Vaswani}, \bibinfo{person}{Noam Shazeer}, \bibinfo{person}{Niki Parmar}, \bibinfo{person}{Jakob Uszkoreit}, \bibinfo{person}{Llion Jones}, \bibinfo{person}{Aidan~N Gomez}, \bibinfo{person}{{\L}ukasz Kaiser}, {and} \bibinfo{person}{Illia Polosukhin}.} \bibinfo{year}{2017}\natexlab{}.
\newblock \showarticletitle{Attention is all you need}.
\newblock \bibinfo{journal}{\emph{Advances in neural information processing systems}}  \bibinfo{volume}{30} (\bibinfo{year}{2017}).
\newblock


\bibitem[Wang et~al\mbox{.}(2021a)]%
        {wang2021survey}
\bibfield{author}{\bibinfo{person}{Sheng Wang}, \bibinfo{person}{Zhifeng Bao}, \bibinfo{person}{J~Shane Culpepper}, {and} \bibinfo{person}{Gao Cong}.} \bibinfo{year}{2021}\natexlab{a}.
\newblock \showarticletitle{A survey on trajectory data management, analytics, and learning}.
\newblock \bibinfo{journal}{\emph{ACM Computing Surveys (CSUR)}} \bibinfo{volume}{54}, \bibinfo{number}{2} (\bibinfo{year}{2021}), \bibinfo{pages}{1--36}.
\newblock


\bibitem[Wang et~al\mbox{.}(2021b)]%
        {wang2021large}
\bibfield{author}{\bibinfo{person}{Xingrui Wang}, \bibinfo{person}{Xinyu Liu}, \bibinfo{person}{Ziteng Lu}, {and} \bibinfo{person}{Hanfang Yang}.} \bibinfo{year}{2021}\natexlab{b}.
\newblock \showarticletitle{Large scale GPS trajectory generation using map based on two stage GAN}.
\newblock \bibinfo{journal}{\emph{Journal of Data Science}} \bibinfo{volume}{19}, \bibinfo{number}{1} (\bibinfo{year}{2021}), \bibinfo{pages}{126--141}.
\newblock


\bibitem[Wang et~al\mbox{.}(2018)]%
        {WDR}
\bibfield{author}{\bibinfo{person}{Zheng Wang}, \bibinfo{person}{Kun Fu}, {and} \bibinfo{person}{Jieping Ye}.} \bibinfo{year}{2018}\natexlab{}.
\newblock \showarticletitle{Learning to Estimate the Travel Time}. In \bibinfo{booktitle}{\emph{Proceedings of the 24th ACM SIGKDD International Conference on Knowledge Discovery \& Data Mining}}. \bibinfo{publisher}{Association for Computing Machinery}, \bibinfo{address}{London, United Kingdom}, \bibinfo{pages}{858–866}.
\newblock


\bibitem[Wang et~al\mbox{.}(2021c)]%
        {wang2021trajectory}
\bibfield{author}{\bibinfo{person}{Zheng Wang}, \bibinfo{person}{Cheng Long}, {and} \bibinfo{person}{Gao Cong}.} \bibinfo{year}{2021}\natexlab{c}.
\newblock \showarticletitle{Trajectory simplification with reinforcement learning}. In \bibinfo{booktitle}{\emph{2021 IEEE 37th International Conference on Data Engineering (ICDE)}}. IEEE, \bibinfo{pages}{684--695}.
\newblock


\bibitem[Wen et~al\mbox{.}(2023)]%
        {wen2023diffstg}
\bibfield{author}{\bibinfo{person}{Haomin Wen}, \bibinfo{person}{Youfang Lin}, \bibinfo{person}{Yutong Xia}, \bibinfo{person}{Huaiyu Wan}, \bibinfo{person}{Qingsong Wen}, \bibinfo{person}{Roger Zimmermann}, {and} \bibinfo{person}{Yuxuan Liang}.} \bibinfo{year}{2023}\natexlab{}.
\newblock \showarticletitle{DiffSTG: Probabilistic Spatio-Temporal Graph Forecasting with Denoising Diffusion Models}.
\newblock \bibinfo{journal}{\emph{arXiv preprint arXiv:2301.13629}} (\bibinfo{year}{2023}).
\newblock


\bibitem[Wu et~al\mbox{.}(2020)]%
        {MTGNN}
\bibfield{author}{\bibinfo{person}{Zonghan Wu}, \bibinfo{person}{Shirui Pan}, \bibinfo{person}{Guodong Long}, \bibinfo{person}{Jing Jiang}, \bibinfo{person}{Xiaojun Chang}, {and} \bibinfo{person}{Chengqi Zhang}.} \bibinfo{year}{2020}\natexlab{}.
\newblock \showarticletitle{Connecting the dots: Multivariate time series forecasting with graph neural networks}. In \bibinfo{booktitle}{\emph{Proceedings of the 26th ACM SIGKDD international conference on knowledge discovery \& data mining}}. \bibinfo{pages}{753--763}.
\newblock


\bibitem[Wu et~al\mbox{.}(2019)]%
        {GWNet}
\bibfield{author}{\bibinfo{person}{Zonghan Wu}, \bibinfo{person}{Shirui Pan}, \bibinfo{person}{Guodong Long}, \bibinfo{person}{Jing Jiang}, {and} \bibinfo{person}{Chengqi Zhang}.} \bibinfo{year}{2019}\natexlab{}.
\newblock \showarticletitle{Graph Wavenet for Deep Spatial-Temporal Graph Modeling}. In \bibinfo{booktitle}{\emph{Proceedings of the 28th International Joint Conference on Artificial Intelligence}} (Macao, China) \emph{(\bibinfo{series}{IJCAI'19})}. \bibinfo{pages}{1907–1913}.
\newblock


\bibitem[Xi et~al\mbox{.}(2018)]%
        {xi2018trajgans}
\bibfield{author}{\bibinfo{person}{Liu Xi}, \bibinfo{person}{Chen Hanzhou}, {and} \bibinfo{person}{Andris Clio}.} \bibinfo{year}{2018}\natexlab{}.
\newblock \showarticletitle{trajGANs: using generative adversarial networks for geo-privacy protection of trajectory data}.
\newblock \bibinfo{journal}{\emph{Vision paper}} (\bibinfo{year}{2018}).
\newblock


\bibitem[Xia et~al\mbox{.}(2018)]%
        {Xia2018deeprailway}
\bibfield{author}{\bibinfo{person}{Tianqi Xia}, \bibinfo{person}{Xuan Song}, \bibinfo{person}{Zipei Fan}, \bibinfo{person}{Hiroshi Kanasugi}, \bibinfo{person}{QuanJun Chen}, \bibinfo{person}{Renhe Jiang}, {and} \bibinfo{person}{Ryosuke Shibasaki}.} \bibinfo{year}{2018}\natexlab{}.
\newblock \showarticletitle{DeepRailway: A Deep Learning System for Forecasting Railway Traffic}. In \bibinfo{booktitle}{\emph{2018 IEEE Conference on Multimedia Information Processing and Retrieval (MIPR)}}. \bibinfo{pages}{51--56}.
\newblock


\bibitem[Xu et~al\mbox{.}(2021)]%
        {xu2021simulating}
\bibfield{author}{\bibinfo{person}{Nan Xu}, \bibinfo{person}{Loc Trinh}, \bibinfo{person}{Sirisha Rambhatla}, \bibinfo{person}{Zhen Zeng}, \bibinfo{person}{Jiahao Chen}, \bibinfo{person}{Samuel Assefa}, {and} \bibinfo{person}{Yan Liu}.} \bibinfo{year}{2021}\natexlab{}.
\newblock \showarticletitle{Simulating continuous-time human mobility trajectories}. In \bibinfo{booktitle}{\emph{Proc. 9th Int. Conf. Learn. Represent}}. \bibinfo{pages}{1--9}.
\newblock


\bibitem[Xu et~al\mbox{.}(2022)]%
        {xu2022metaptp}
\bibfield{author}{\bibinfo{person}{Yuan Xu}, \bibinfo{person}{Jiajie Xu}, \bibinfo{person}{Jing Zhao}, \bibinfo{person}{Kai Zheng}, \bibinfo{person}{An Liu}, \bibinfo{person}{Lei Zhao}, {and} \bibinfo{person}{Xiaofang Zhou}.} \bibinfo{year}{2022}\natexlab{}.
\newblock \showarticletitle{MetaPTP: an adaptive meta-optimized model for personalized spatial trajectory prediction}. In \bibinfo{booktitle}{\emph{Proceedings of the 28th ACM SIGKDD Conference on Knowledge Discovery and Data Mining}}. \bibinfo{pages}{2151--2159}.
\newblock


\bibitem[Yang et~al\mbox{.}(2022)]%
        {diffusionsuvery}
\bibfield{author}{\bibinfo{person}{Ling Yang}, \bibinfo{person}{Zhilong Zhang}, \bibinfo{person}{Yang Song}, \bibinfo{person}{Shenda Hong}, \bibinfo{person}{Runsheng Xu}, \bibinfo{person}{Yue Zhao}, \bibinfo{person}{Yingxia Shao}, \bibinfo{person}{Wentao Zhang}, \bibinfo{person}{Bin Cui}, {and} \bibinfo{person}{Ming-Hsuan Yang}.} \bibinfo{year}{2022}\natexlab{}.
\newblock \showarticletitle{Diffusion models: A comprehensive survey of methods and applications}.
\newblock \bibinfo{journal}{\emph{arXiv preprint arXiv:2209.00796}} (\bibinfo{year}{2022}).
\newblock


\bibitem[Yuan et~al\mbox{.}(2022)]%
        {yuan2022activity}
\bibfield{author}{\bibinfo{person}{Yuan Yuan}, \bibinfo{person}{Jingtao Ding}, \bibinfo{person}{Huandong Wang}, \bibinfo{person}{Depeng Jin}, {and} \bibinfo{person}{Yong Li}.} \bibinfo{year}{2022}\natexlab{}.
\newblock \showarticletitle{Activity Trajectory Generation via Modeling Spatiotemporal Dynamics}. In \bibinfo{booktitle}{\emph{Proceedings of the 28th ACM SIGKDD Conference on Knowledge Discovery and Data Mining}}. \bibinfo{pages}{4752--4762}.
\newblock


\bibitem[Zandbergen(2014)]%
        {zandbergen2014ensuring}
\bibfield{author}{\bibinfo{person}{Paul~A Zandbergen}.} \bibinfo{year}{2014}\natexlab{}.
\newblock \showarticletitle{Ensuring confidentiality of geocoded health data: assessing geographic masking strategies for individual-level data}.
\newblock \bibinfo{journal}{\emph{Advances in medicine}}  \bibinfo{volume}{2014} (\bibinfo{year}{2014}).
\newblock


\bibitem[Zhang et~al\mbox{.}(2022)]%
        {zhang2022dp}
\bibfield{author}{\bibinfo{person}{Jing Zhang}, \bibinfo{person}{Qihan Huang}, \bibinfo{person}{Yirui Huang}, \bibinfo{person}{Qian Ding}, {and} \bibinfo{person}{Pei-Wei Tsai}.} \bibinfo{year}{2022}\natexlab{}.
\newblock \showarticletitle{DP-TrajGAN: A privacy-aware trajectory generation model with differential privacy}.
\newblock \bibinfo{journal}{\emph{Future Generation Computer Systems}} (\bibinfo{year}{2022}).
\newblock


\bibitem[Zheng(2015)]%
        {zheng2015trajectory}
\bibfield{author}{\bibinfo{person}{Yu Zheng}.} \bibinfo{year}{2015}\natexlab{}.
\newblock \showarticletitle{Trajectory data mining: an overview}.
\newblock \bibinfo{journal}{\emph{ACM Transactions on Intelligent Systems and Technology (TIST)}} \bibinfo{volume}{6}, \bibinfo{number}{3} (\bibinfo{year}{2015}), \bibinfo{pages}{1--41}.
\newblock


\bibitem[Zheng et~al\mbox{.}(2014)]%
        {zheng2014urban}
\bibfield{author}{\bibinfo{person}{Yu Zheng}, \bibinfo{person}{Licia Capra}, \bibinfo{person}{Ouri Wolfson}, {and} \bibinfo{person}{Hai Yang}.} \bibinfo{year}{2014}\natexlab{}.
\newblock \showarticletitle{Urban computing: concepts, methodologies, and applications}.
\newblock \bibinfo{journal}{\emph{ACM Transactions on Intelligent Systems and Technology (TIST)}} \bibinfo{volume}{5}, \bibinfo{number}{3} (\bibinfo{year}{2014}), \bibinfo{pages}{1--55}.
\newblock


\bibitem[Zhu et~al\mbox{.}(2021)]%
        {zhu2021semi}
\bibfield{author}{\bibinfo{person}{Yuanshao Zhu}, \bibinfo{person}{Yi Liu}, \bibinfo{person}{James J.~Q. Yu}, {and} \bibinfo{person}{Xingliang Yuan}.} \bibinfo{year}{2021}\natexlab{}.
\newblock \showarticletitle{Semi-supervised federated learning for travel mode identification from gps trajectories}.
\newblock \bibinfo{journal}{\emph{IEEE Transactions on Intelligent Transportation Systems}} \bibinfo{volume}{23}, \bibinfo{number}{3} (\bibinfo{year}{2021}), \bibinfo{pages}{2380--2391}.
\newblock


\bibitem[Zhu et~al\mbox{.}(2023)]%
        {zhu2023difftraj}
\bibfield{author}{\bibinfo{person}{Yuanshao Zhu}, \bibinfo{person}{Yongchao Ye}, \bibinfo{person}{Shiyao Zhang}, \bibinfo{person}{Xiangyu Zhao}, {and} \bibinfo{person}{James~Jianqiao Yu}.} \bibinfo{year}{2023}\natexlab{}.
\newblock \showarticletitle{DiffTraj: Generating {GPS} Trajectory with Diffusion Probabilistic Model}. In \bibinfo{booktitle}{\emph{Thirty-seventh Conference on Neural Information Processing Systems}}.
\newblock


\end{thebibliography}

\clearpage

\appendix
\begin{appendix}
\begin{figure*}[htbp]
    \includegraphics[width=0.9\linewidth]{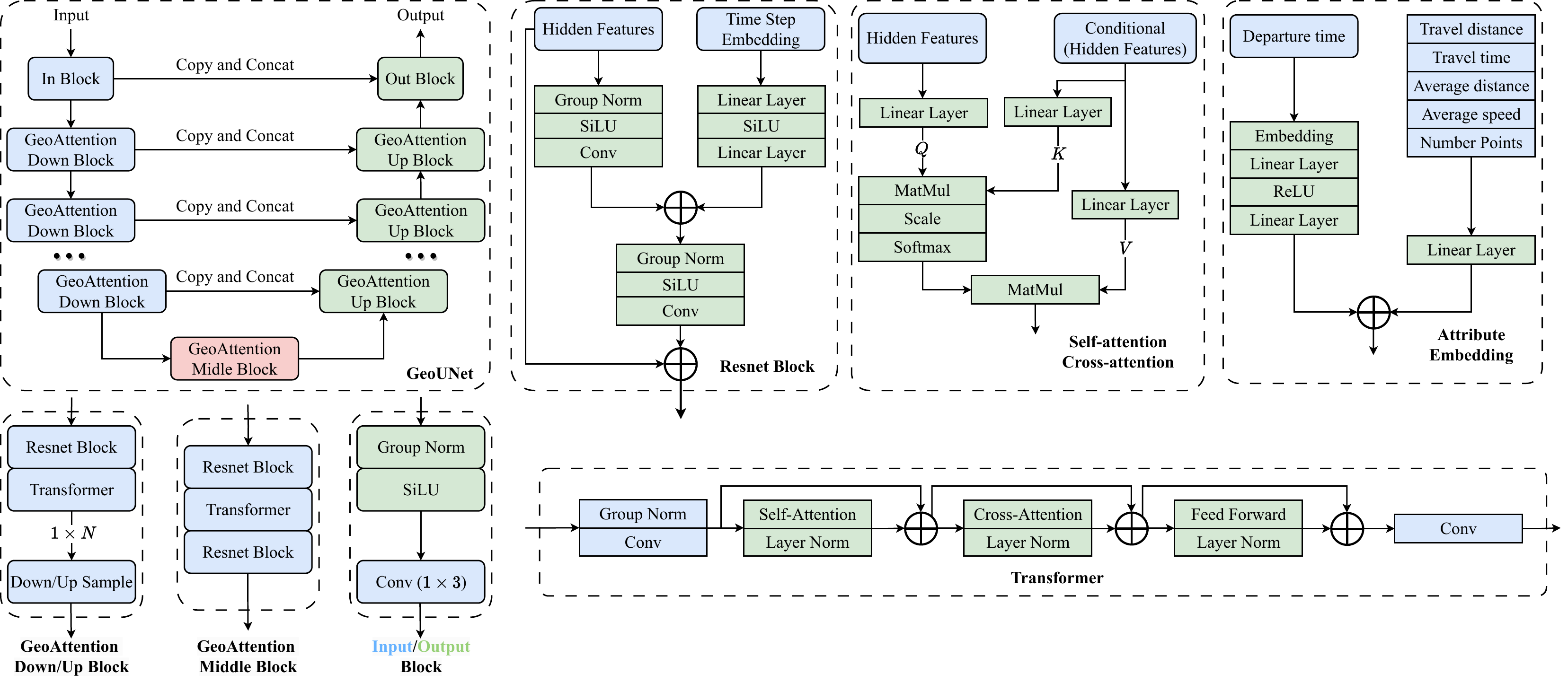}
    \caption{The main architecture and compoents of GeoUNet.}
    \label{fig:geounet}
\end{figure*}

\section{Details of ControlTraj Framework}

In this section, we provide a detailed implementation of ControlTraj, including the GeoUNet architecture, parameter settings.

\subsection{GeoUNet architecture}\label{app:geounet}

As illustrated in \cref{fig:geounet}, ControlTraj is divided into two parts, downsampling and upsampling, and each part consists of multiple stacked Geo Down/Up Blocks with integrated geographic attention and residual modules.
To learn road segment topological constraints and form attributes, ControlTraj integrates an attribute module to embed attribute information, which is then concatenated with the road segment embedding and collectively fed to each block.
Specifically, each sampling module consists of a collection of components such as residual blocks, transformers, etc.
Among them, for downsampling the max pooling layer is used and linear interpolation is applied for upsampling.
For trip attribute information, such as distance, speed, departure time, travel time, etc., we follow the practice in \cite{WDR,zhu2023difftraj} and integrate a Wide and Deep module.
In addition, road topology constraints are incorporated into GeoUNet through transformer layers, each of which includes a series of group normalization, attention layer, and fully connected layers. 
Notably, two cascaded attention mechanisms are included here, i.e., self-attention and cross-attention.

\subsection{Implementation Details}\label{app:imple}

\begin{table}[h]
    \caption{General setting for RoadMAE.}
    \centering
    \begin{tabular}{lcrr} 
    \toprule
    Parameter & & Setting value & Refer range  \\ 
    \cmidrule(lr){1-4}
    Mask ratio & & 0.5  & 0 $\sim$ 0.75\\
    Patch length & & 5 & 5 $\sim$ 10\\
    Encoder Blocks & & 8 & $\ge$ 2\\
    Decoder Blocks & & 4 & $\ge$ 2\\
    Heads & & 4 & $\ge$ 1\\
    Encode Dim & & 128 & $\ge$ 64 \\
    Parameter size & & 9.2 MB  & -- \\
     \bottomrule
    \end{tabular}
    \label{tab:roadmae_para}
\end{table}

\begin{table}[h]
    \caption{General setting for GeoUNet.}
    \centering
    \begin{tabular}{lcrr} 
    \toprule
    Parameter & & Setting value & Refer range  \\ 
    \cmidrule(lr){1-4}
    Diffusion Steps & & 500  & 300 $\sim$ 500\\
    Skip steps & & 5 & 1 $\sim$ 10\\
    Embedding dim & & 128 & $\ge$ 64 \\
    $\beta$ (linear schedule) & & 0.0001 $\sim$ 0.05 & -- \\
    Batch size & & 1024 & $\ge$ 256 \\
    Sampling blocks & & 4 &  $\ge$ 3 \\
    Resnet blocks & & 2 & $\ge$ 1\\
    Input Length & & 200 & 120 $\sim$ 256\\
    Parameter size & & 31.5 MB  & -- \\
     \bottomrule
    \end{tabular}
    \label{tab:model_para}
\end{table}

In addition, we summarize the list of hyperparameters and the specific implementation settings used in this paper.
Specifically, we summarize the parameter settings for RoadMAE in \cref{tab:roadmae_para} and GeoUNet in \cref{tab:model_para}.
Meanwhile, we provide reference ranges for the settings of these hyperparameters based on the results of the current work and general practical experience.

\section{Experiment and Setup}\label{app:exp_setup}
In this section, we describe the experimental setup of this paper in detail, including the dataset, baseline implementation and evaluation metrics.
Please kindly note that all experiments are implemented with PyTorch 3.9 and conducted on a single NVIDIA A100 40GB GPU.

\begin{figure*}[t]
    \subfigure[Xi'an]{
    \includegraphics[width=0.32\linewidth]{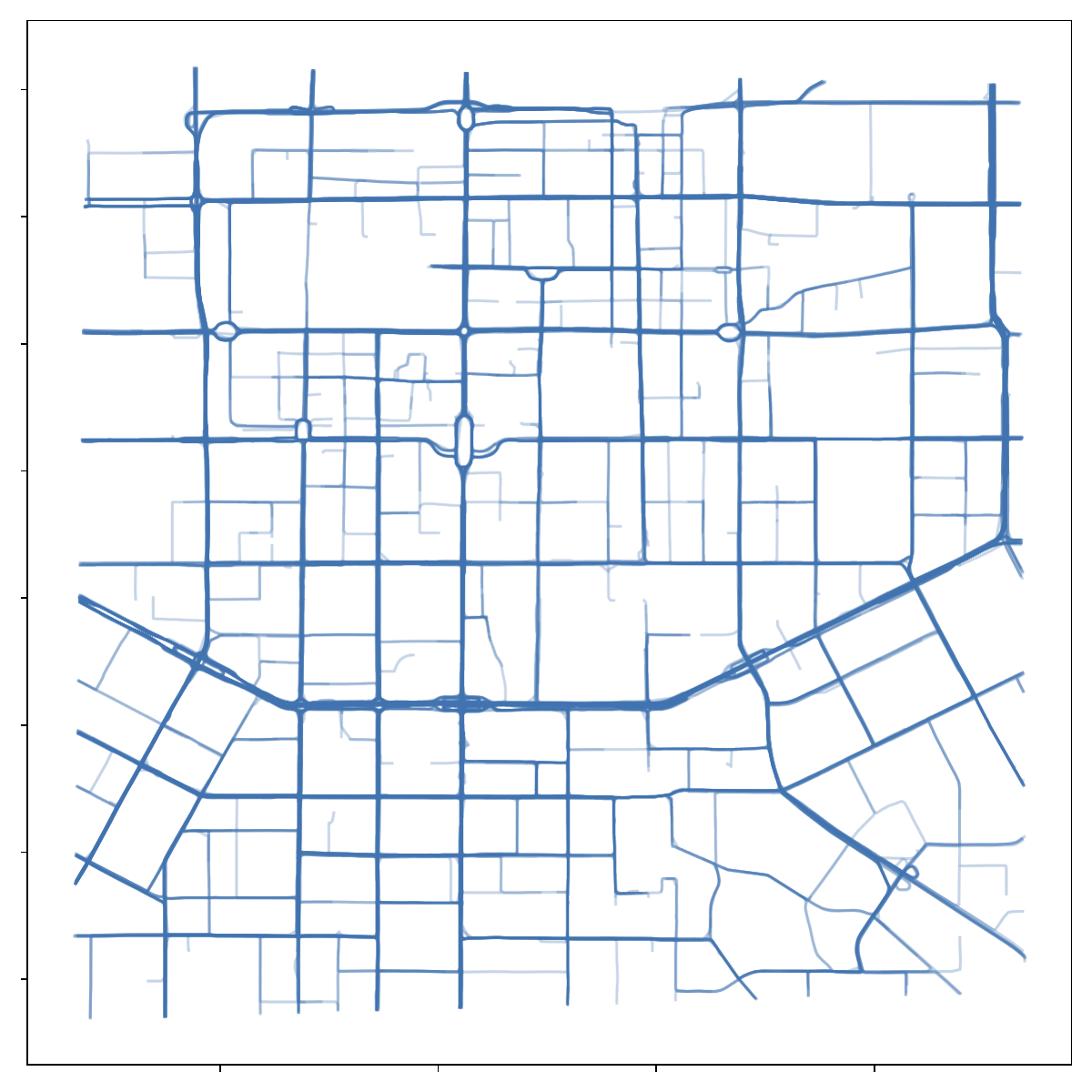} 
    }
    \subfigure[Chengdu]{
    \includegraphics[width=0.32\linewidth]{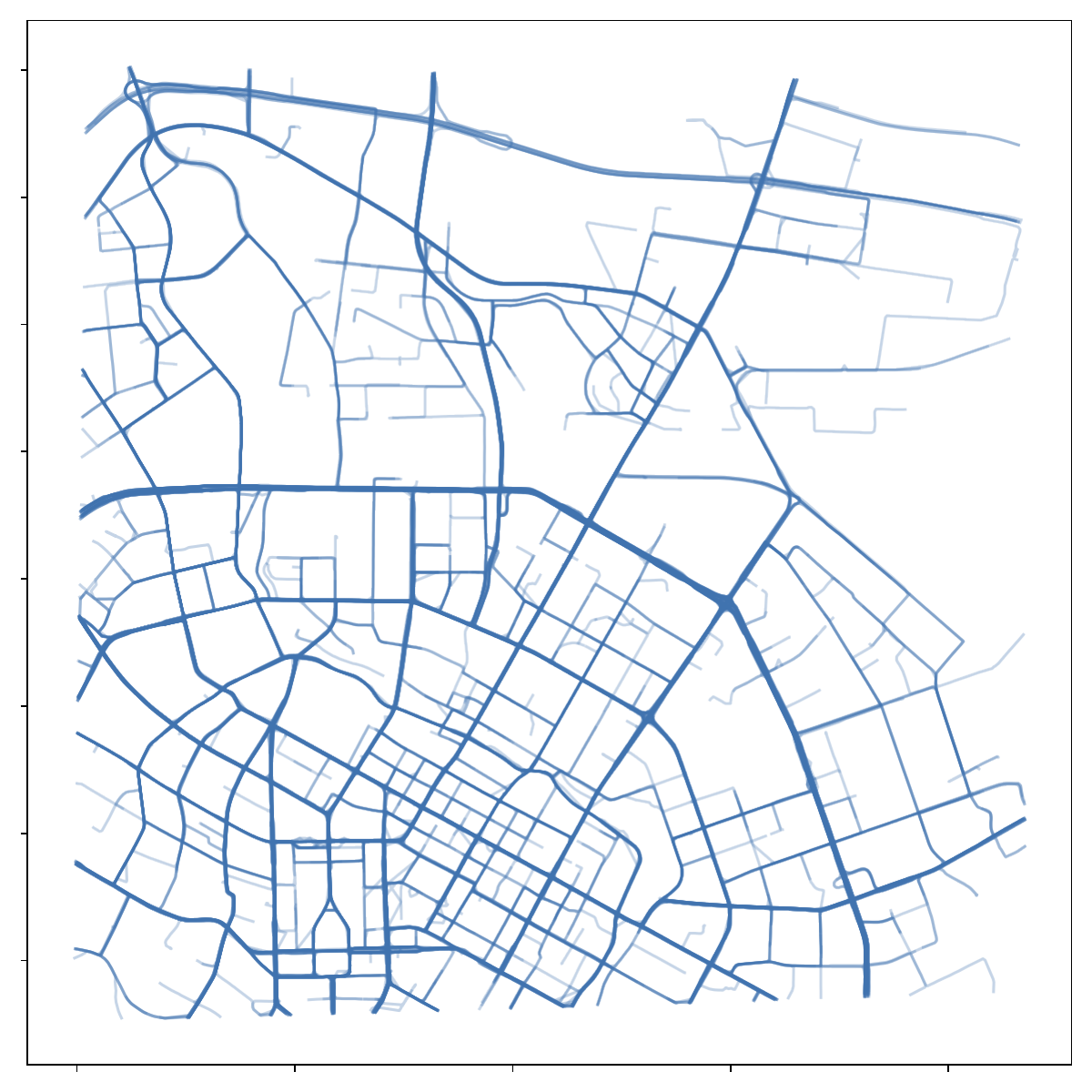}
    }
    \subfigure[Porto]{
    \includegraphics[width=0.32\linewidth]{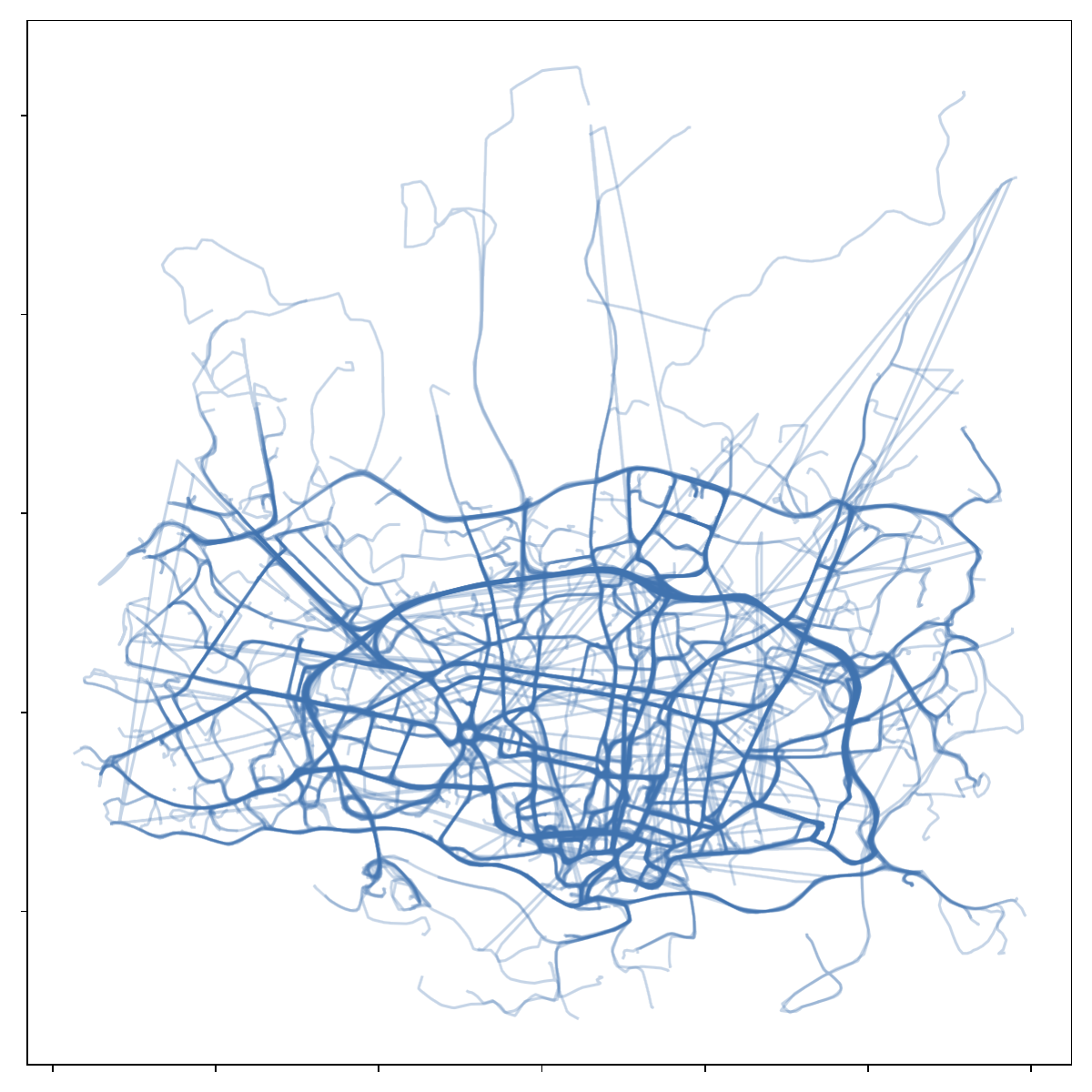}
    }
    \caption{Origin trajectory distribution of three cities.}
    \label{fig:original}
\end{figure*}

\subsection{Dataset}\label{app:dataset}
We evaluate the model performance of \model and baselines on three datasets with different cities, \textbf{Chengdu}, \textbf{Xi'an}\footnote{https://outreach.didichuxing.com/}, and \textbf{Porto}\footnote{https://www.kaggle.com/datasets/crailtap/taxi-trajectory/}.
\cref{tab:dataset}  shows the statistic of these datasets and \cref{fig:original} show the distribution, respectively.
Deeper colors indicate more frequent trajectories in the region, and lighter ones indicate sparse trajectories.
For all datasets, we filtered trajectories with lengths less than $120$ and normalized them to a fixed length using linear interpolation.

\begin{table}[h]
    \caption{Statistics of Real-world Trajectory Datasets.}
    \centering
    \begin{tabular}{lccc} 
    \toprule
    Dataset & Trajetory Number  & Avg. Length & Avg. Distance    \\ 
    \cmidrule(lr){1-4}
    Chengdu  & \num{5773525}  & 175.9  &  \SI{7.42}{\kilo \meter} \\
    Xian  & \num{3044828}  &  243.8  & \SI{5.73}{\kilo \meter} \\
    Porto  & \num{1710670}  & 48.9 & \SI{4.12}{\kilo \meter} \\
    \bottomrule
    \end{tabular}
    \label{tab:dataset}
\end{table}


\subsection{Baselines}\label{app:baseline}

\begin{itemize}[leftmargin=*] 

\item \textbf{VAE} \cite{Xia2018deeprailway}
We initially encode trajectory data into a concealed distribution using a sequence of two convolutional layers followed by a linear layer. Subsequent to this encoding, the trajectory generation is accomplished through a decoder that incorporates a linear layer and two deconvolutional layers. To maintain consistency in the dimensions of both input and output trajectories, we establish the dimensionality of the convolution and deconvolution kernels at $4$.

\item \textbf{TrajGAN} \cite{xi2018trajgans} 
Here, the trajectory data is amalgamated with stochastic noise before being processed through a generator, which is composed of two linear layers and two convolution layers. 
Following this, the discrimination process utilizes a convolutional layer coupled with a linear layer. Training of the generator and discriminator proceeds in a sequential, alternating fashion, optimizing their performance iteratively.

\item \textbf{Dp-TrajGAN} \cite{zhang2022dp, rao2020lstm}
Since adding differential privacy affects the performance of the model, here we remove the differential privacy operation. Therefore, the model is practically very similar to TrajGAN, with the difference that the backbone layer of the model is an LSTM instead of a CNN.

\item \textbf{Diffwave} \cite{kongdiffwave}
This model is built upon the Wavenet architecture, and is tailored for synthesizing sequences like timeseries, and speech voice. 
This model makes significant use of dilated convolution techniques. 
In our implementation, we incorporate $16$ blocks connected through residual links, where each block features a bidirectional dilated convolution. 
The outputs from these blocks are integrated using sigmoid and tanh activation functions, respectively, before being processed by a one-dimensional Convolutional Neural Network (1D CNN).

\item \textbf{DiffTraj} \cite{zhu2023difftraj}
DiffTraj is an advanced trajectory generation framework based on diffusion models. Here, we follow the model and setup in code repository \footnote{https://github.com/Yasoz/DiffTraj}. 
Note that since the diffTraj does not feature topological guidance for road segments, we simply embed the regions to which the start and end areas belong as conditions.

\item \textbf{\model w/o $\boldsymbol{c}$}
This baseline has the same settings and GeoUNet structure as ControlTraj. The difference is that this baseline removes the section topology constraints and attribute information. It can be used to analyze the effect of condition information on ControlTraj.

\item \textbf{\model - AE} 
This baseline replaces the road network topology encoding in ControlTraj, RoadMAE with a valina Autoencoder.
It is used to explore the learning ability of the proposed RoadMAE for road topology constraints.
Finally, the encoder embeds the road segment topology in the same dimension as RoadMAE

\end{itemize}

\subsection{Evaluation Metrics}\label{app:metrics}
The objective of trajectory generation is to create trajectories that mimic real-world movements to support and enhance downstream applications. 
To assess the degree of resemblance between the generated and actual trajectories, it is crucial to evaluate their "similarity". In alignment with methodologies adopted in prior research, such as highlighted by Du et al. (2023) \cite{du2023ldptrace,zhu2023difftraj}, we utilize the Jenson-Shannon divergence (JSD) to gauge the quality of the trajectories produced. JSD serves as a metric for comparing the distributions of real and synthesized trajectories, where a reduced JSD value signifies a closer approximation to the original statistical characteristics.
Let $P$ represent the probability distribution of the original data and $Q$ denote the probability distribution of the generated data; the JSD is determined by the following formula:
\begin{equation}
    \operatorname{JSD}(P \| Q)=\frac{1}{2} \mathbb{D}(P \| M) + \frac{1}{2} \mathbb{D}(Q \| M),
\end{equation}
where $M = \frac{1}{2}(P+Q)$ is the mixture distribution of $P$ and $Q$.

In the evaluation process, we divide each city into grids of $16 \times 16$ size and count the frequency of trajectory points associated with each grid. Then, we can calculate each evaluation metric based on the calculated matrix.
Specially, we evaluate the performance of different trajectory generation methods with following metrics:

\begin{itemize}[leftmargin=*] 
\item \textbf{Density error:} 
This metric counts the geographic occurrences of all trajectory points in the city and serves as a global metric to assess the similarity of geographic distribution between the entire generated trajectory and the original counterpart.
Assuming that $\mathcal{M}(G)$ is the distribution matrix of the generated trajectories and $\mathcal{M}(P)$ is the distribution of the original trajectories, this evaluation metric is formulated as:
\begin{align}
\text{Density Error} =  \operatorname{JSD}\left(\mathcal{M}(G), \mathcal{M}(P) \right),
\end{align}
where $\operatorname{JSD}(\cdot)$ represents the Jenson-Shannon divergence calculation between two distributions.

\item \textbf{Trip error:}
This metric evaluates the trajectory at a userlevel by analyzing the correlation between the starting and ending points of a travel trajectory. To conduct this assessment, we calculate the probability distribution of the start and end points for both the original and the generated trajectories. The JSD is then employed to measure the similarity between these distributions. By using JSD in this context, we aim to quantify the degree to which the generated trajectories accurately reflect the spatial patterns observed in the original data, with a focus on how closely the start and end points of these trips match.

\item \textbf{Length error:} 
This metric is designed to evaluate the distribution of travel distances at the trajectory level. It involves calculating the geographical distance between consecutive points within a trajectory. 
By analyzing these distances, we can assess the overall travel distance distribution of the trajectories, providing insight into the spatial characteristics and movement patterns encapsulated by the data. This approach allows for a detailed examination of how closely the generated trajectories replicate the travel distances observed in the real-world trajectories they aim to mimic.
Specifically, we compute the distance distributions for each of the two trajectories and then compute the JSD between the two distributions.

\end{itemize}

\section{Supplementary Experiments}
In this section, we provide a detailed presentation and setup of the experimental effects.

\subsection{Visualization}\label{app:geovis}

\begin{figure*}[h]
    \includegraphics[width=0.95\linewidth]{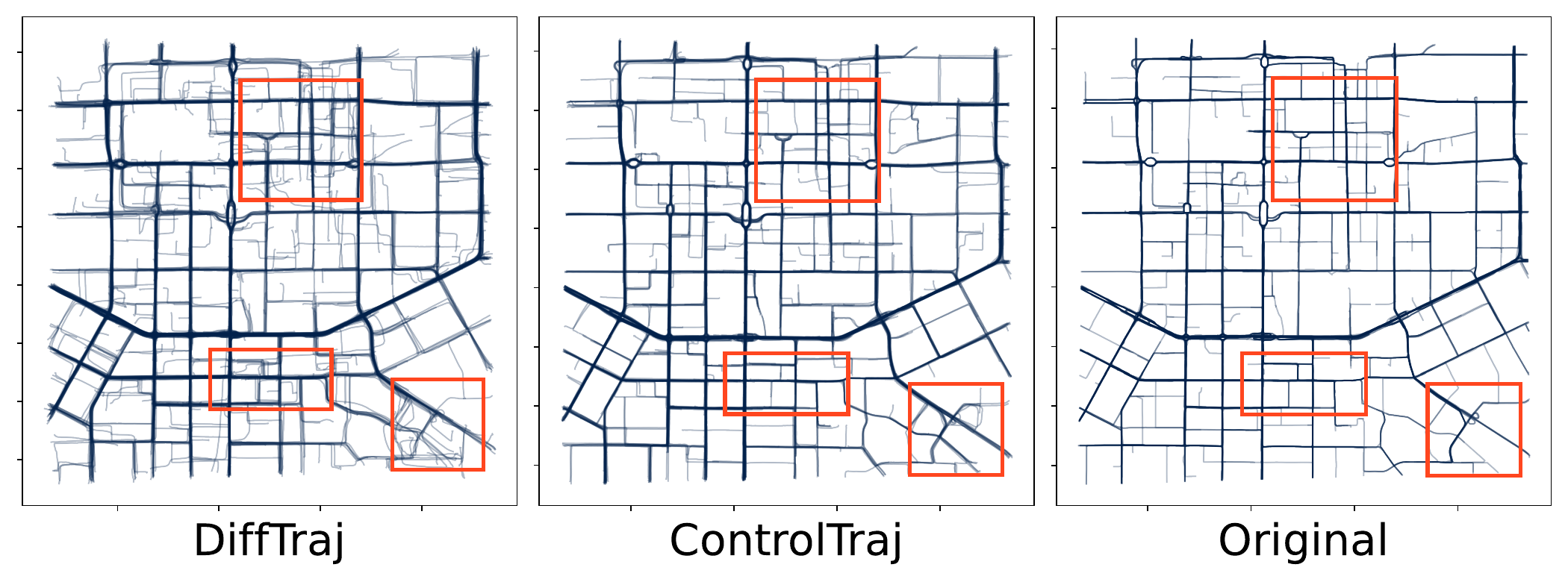}
    \caption{Generated trajectories comparison with Visualization.(Xi'an)}
    \label{fig:app_dis_xa}
\end{figure*}

\begin{figure*}[h]
    \includegraphics[width=0.95\linewidth]{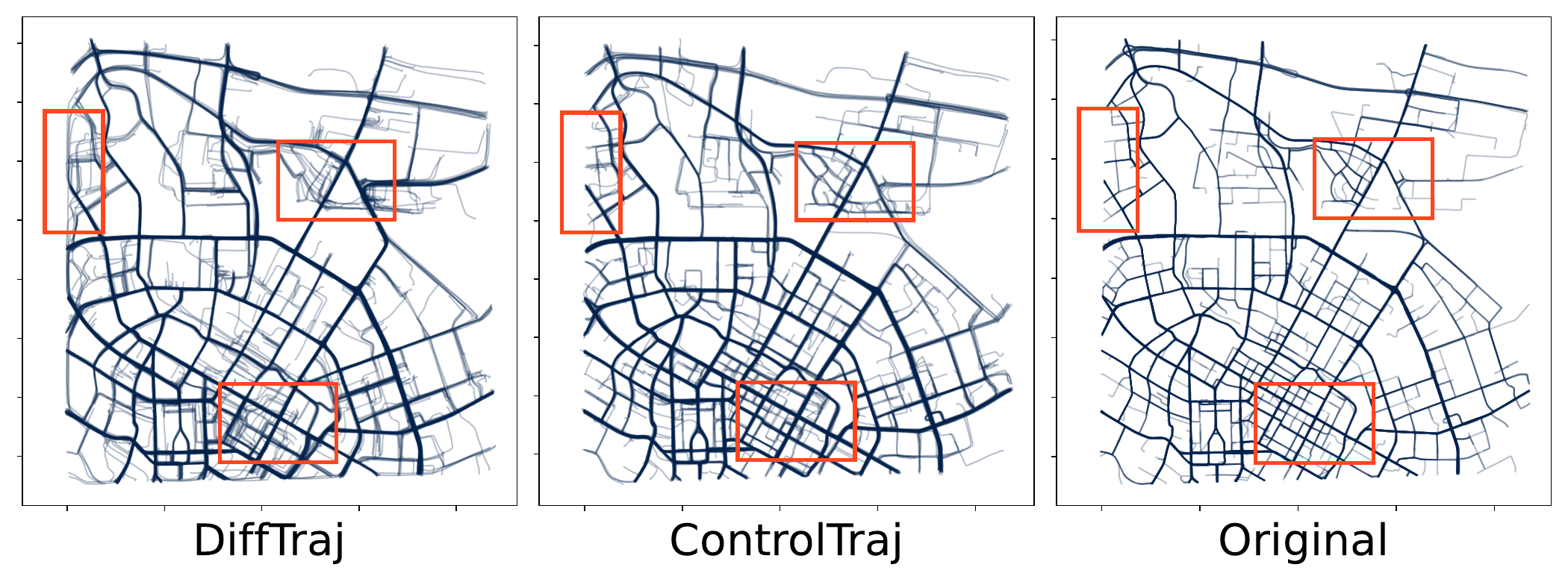}
    \caption{Generated trajectories comparison with Visualization.(Chengdu)}
     \label{fig:app_dis_cd}
\end{figure*}

\begin{figure*}[h]
    \includegraphics[width=0.95\linewidth]{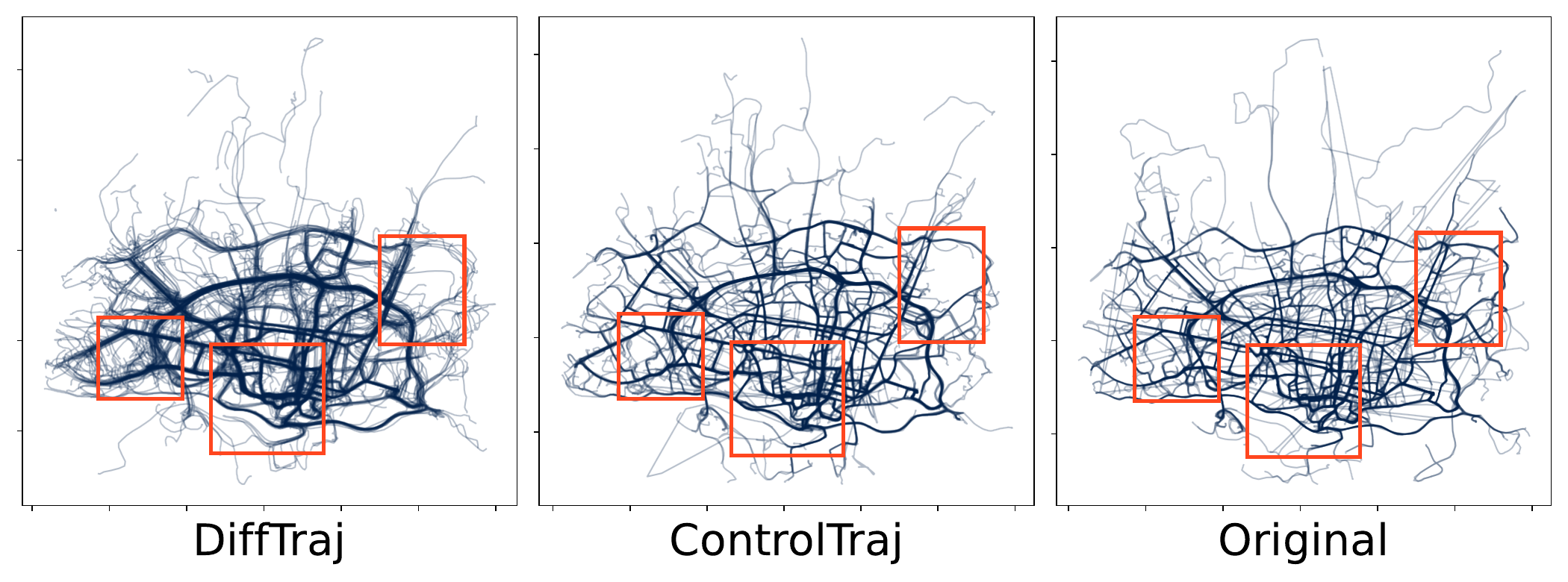}
    \caption{Generated trajectories comparison with Visualization.(Porto)}
     \label{fig:app_dis_porto}
\end{figure*}

Here, we show the trajectories generated by different trajectory generation methods and a large view comparison of the one-on-one geographic results with the original trajectories. In particular, \cref{fig:app_dis_xa} shows Xian, \cref{fig:app_dis_cd} shows Chengdu, and \cref{fig:app_dis_porto} shows Porto. The main differences between the two methods are labeled through red boxes.
From the refined visualization, we can see that the trajectories generated by the \model show higher consistency with the original trajectories, especially for those complex trajectory scenarios, or regions with sparse trajectories.

\subsection{Generated Trajectories Analysis}\label{app:analysis}
\vspace{-2mm}
\begin{table}[h]
\centering
\small
    \caption{Data utility comparison by traffic flow prediction (Xi'an). Results are expressed as (original / generated).}
    \vspace{-3mm}
\begin{tabular}{lcccc}
\toprule
Methods & ASTGCN & GWNet & MTGNN & DCRNN  \\
\cmidrule{1-5}
RMSE    &  5.39 / 4.91  &  6.69 / 6.19 &  6.34 / 5.58  &  6.52 / 5.76 \\
MAE     &  3.26 / 3.00 &  4.53 / 4.03  & 4.29 / 3.82 &  4.48 / 3.97 \\
MAPE    &  23.07 / 23.83 & 29.42 / 34.37  & 28.70 / 30.95  & 32.82 / 32.15   \\
\bottomrule
\end{tabular}
\label{tab:trafficflow_xa}
\vspace{-3mm}
\end{table}

\begin{figure*}[t]
    \subfigure{
        \includegraphics[width=0.48\linewidth]{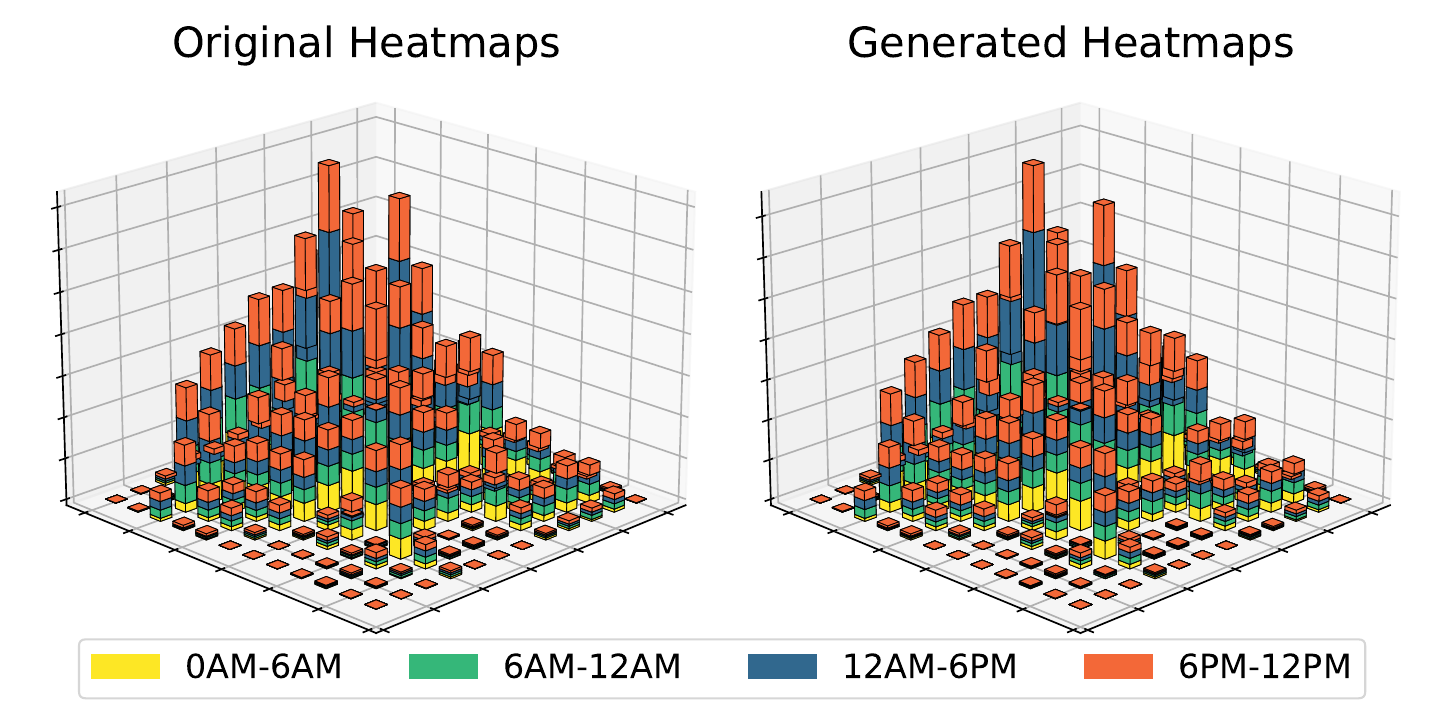}
    }
    \subfigure{
        \includegraphics[width=0.48\linewidth]{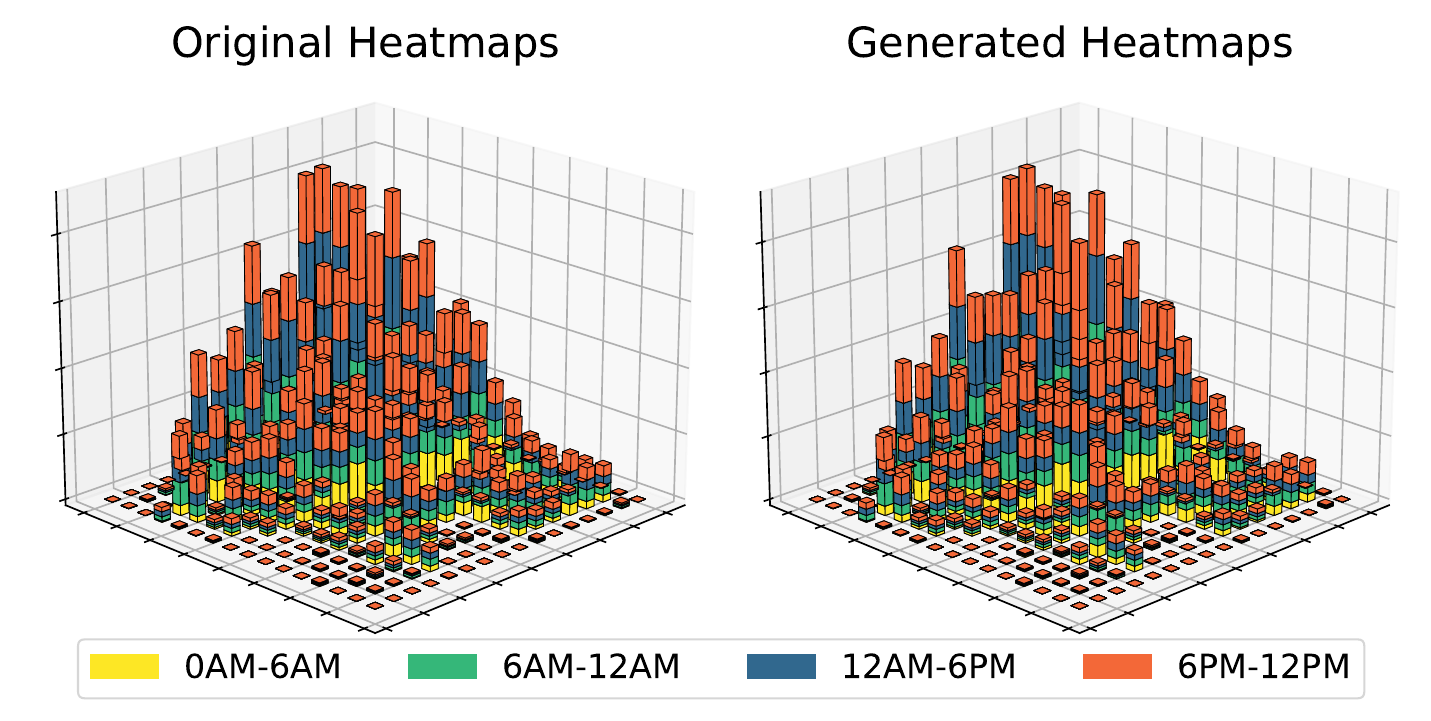}
    }
    \caption{Comparison of heatmaps with generated trajectories (Chengdu).}
    \label{fig:app_heatmap_cd}
\end{figure*}

\begin{figure*}[t]
    \subfigure[Chengdu]{
        \includegraphics[width=0.48\linewidth]{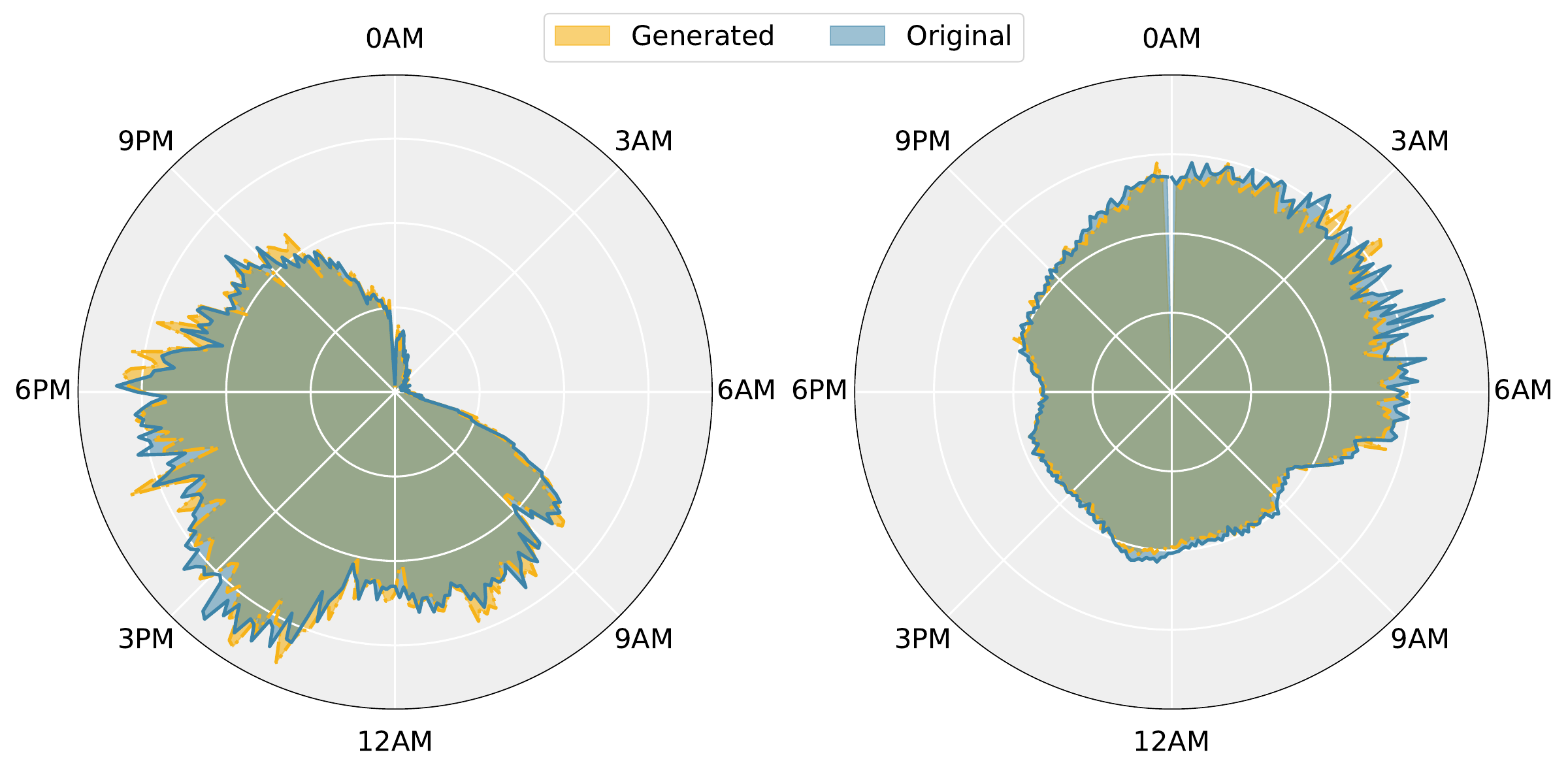}
    }
    \subfigure[Xi'an]{
        \includegraphics[width=0.48\linewidth]{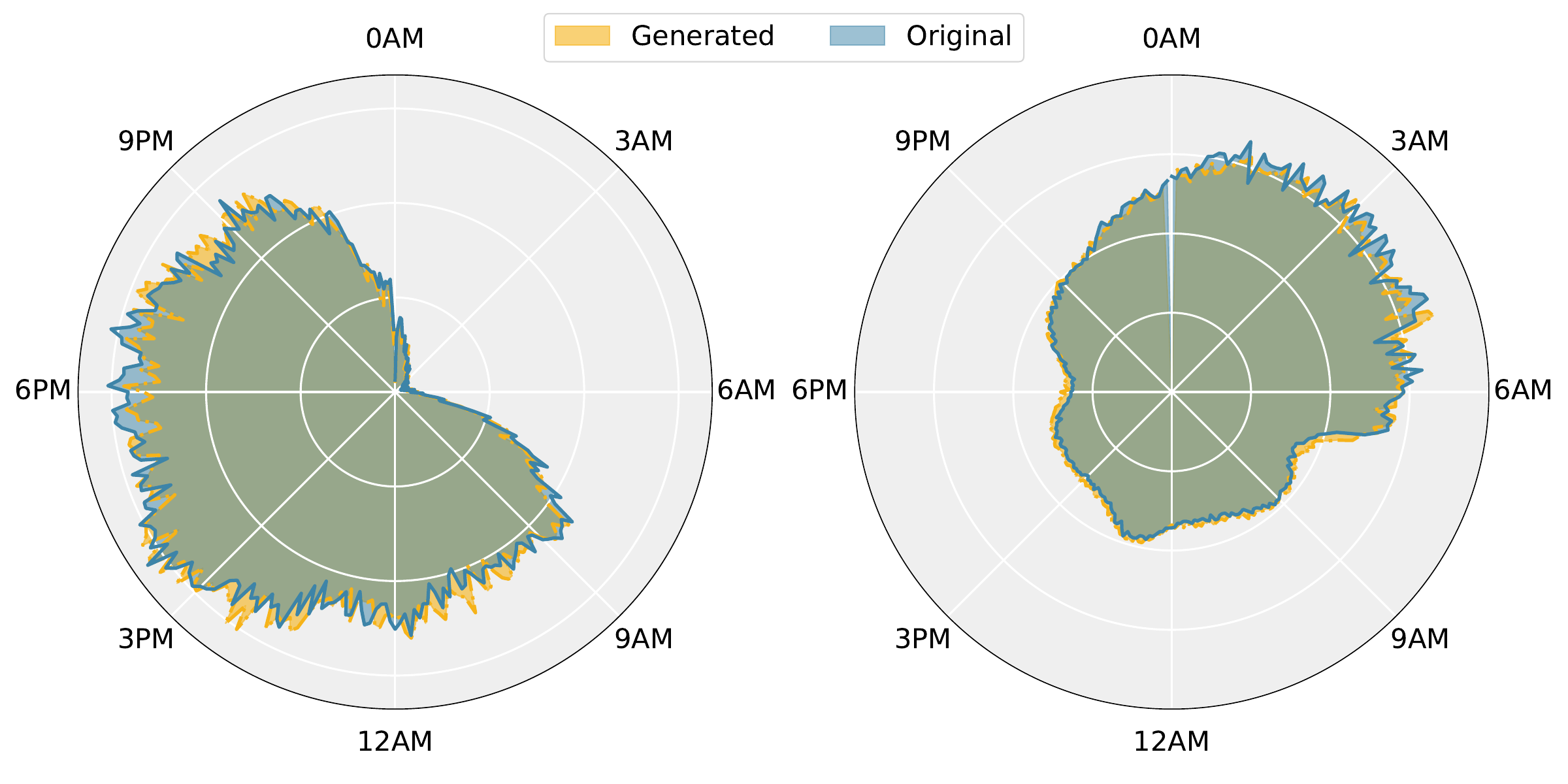}
    }
    \caption{Comparison of volume of trips and average speed with generated trajectories}
    \label{fig:app_radar}
\end{figure*}

In this section, we provide heat maps of the all-day trajectory distribution for the cities of Xi'an (\cref{{fig:app_heatmap_xa}}) and Chengdu (\cref{fig:app_heatmap_cd}). 
In addition, \cref{fig:app_radar} shows the change in trip volumes and average speed over the day.
Note that here we divide the city areas into smaller grids ($12 \times 12$ and $16 \times 16$) to provide a better and more detailed comparison.
In addition, we did the same data availability experiments for the trajectories generated in Xi'an city, and the results are summarized in \cref{tab:trafficflow_xa}.
To evaluate the performance of these tasks and the model, we adopt three metrics:
\begin{align}
     {\rm RMSE} &= \sqrt{\frac{1}{n}\sum_{i}^{n}\left(y_{i}-\hat{y}_{i}\right)^{2}},\\
     {\rm MAE} &= \frac{1}{n}\sum_{i}^{n}\left|{y_{i}-\hat{y}_{i}}\right|,\\
     {\rm MAPE} &= \frac{1}{n}\sum_{i}^{n}\left|\frac{{y_{i}-\hat{y}_{i}}}{y_{i}}\right|,
\end{align}
where $y_{i}$ represents the actual traffic value and  $\hat{y}_{i}$ denotes the predicted traffic value for each observation.

\begin{figure*}[t]
    \subfigure{
        \includegraphics[width=0.48\linewidth]{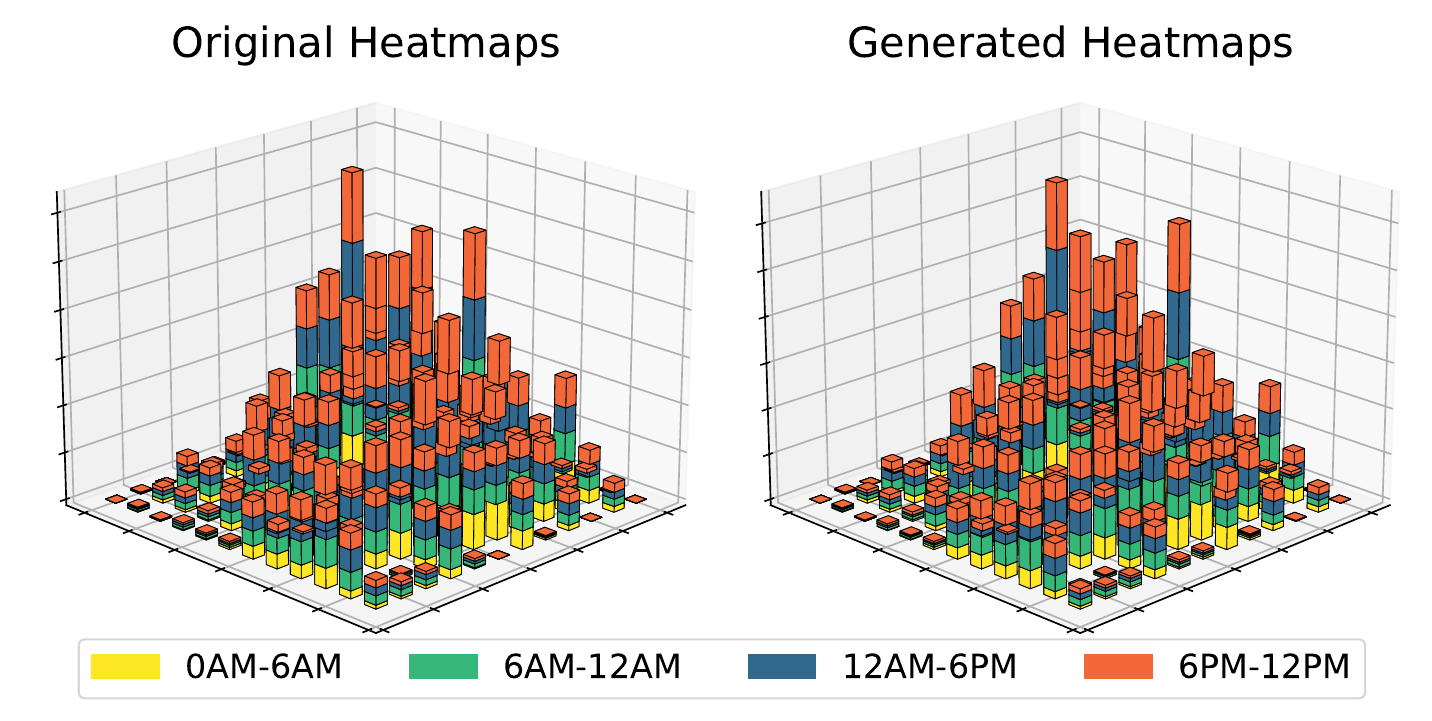}
    }
    \subfigure{
        \includegraphics[width=0.48\linewidth]{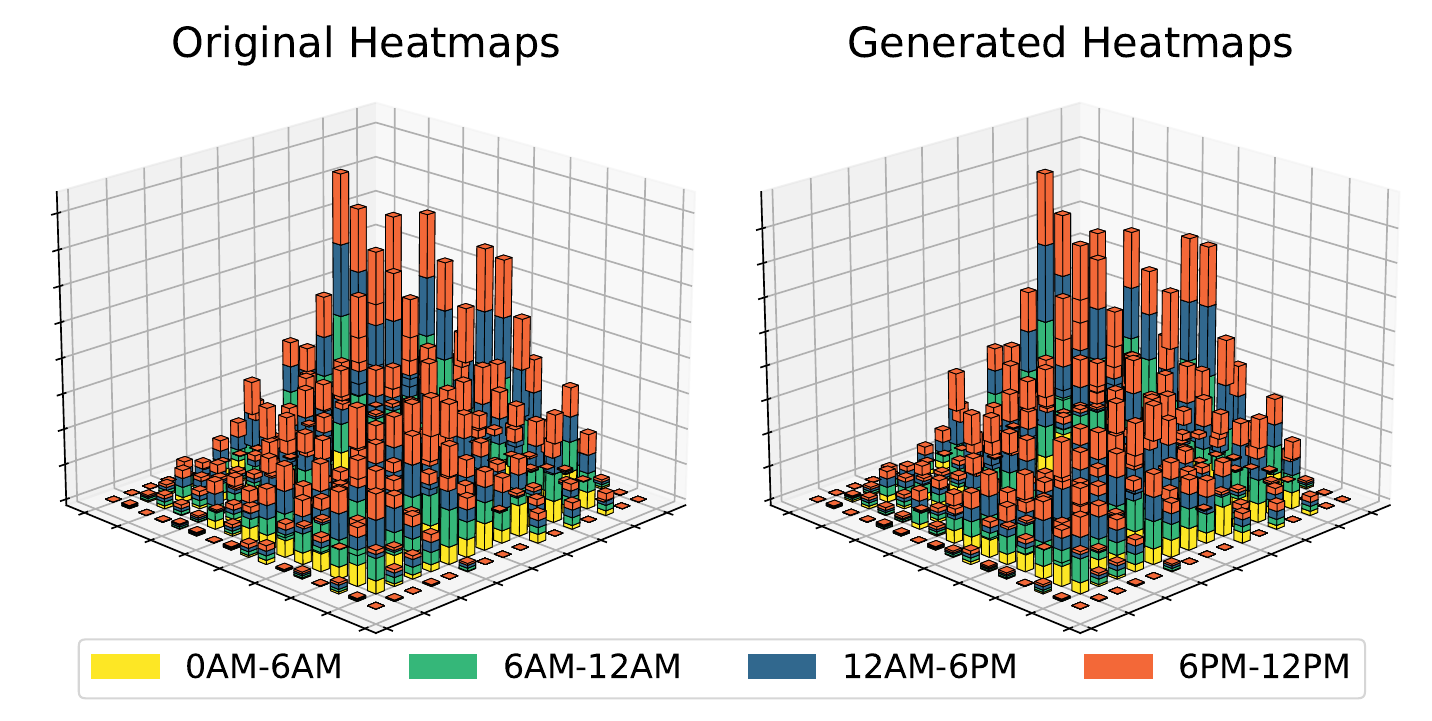}
    }
    \caption{Comparison of heatmaps with generated trajectories (Xi'an).}
    \label{fig:app_heatmap_xa}
\end{figure*}

\begin{figure*}
    \centering
    \subfigure{
        \includegraphics[width=0.48\linewidth]{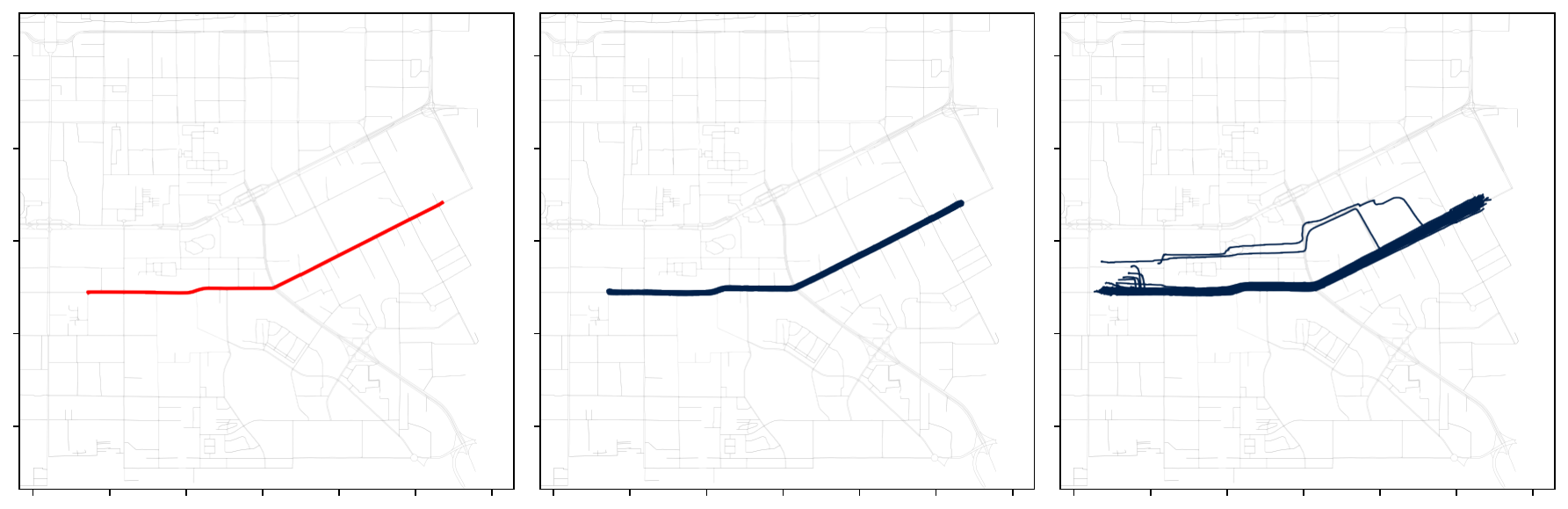}
    }
    \subfigure{
        \includegraphics[width=0.47\linewidth]{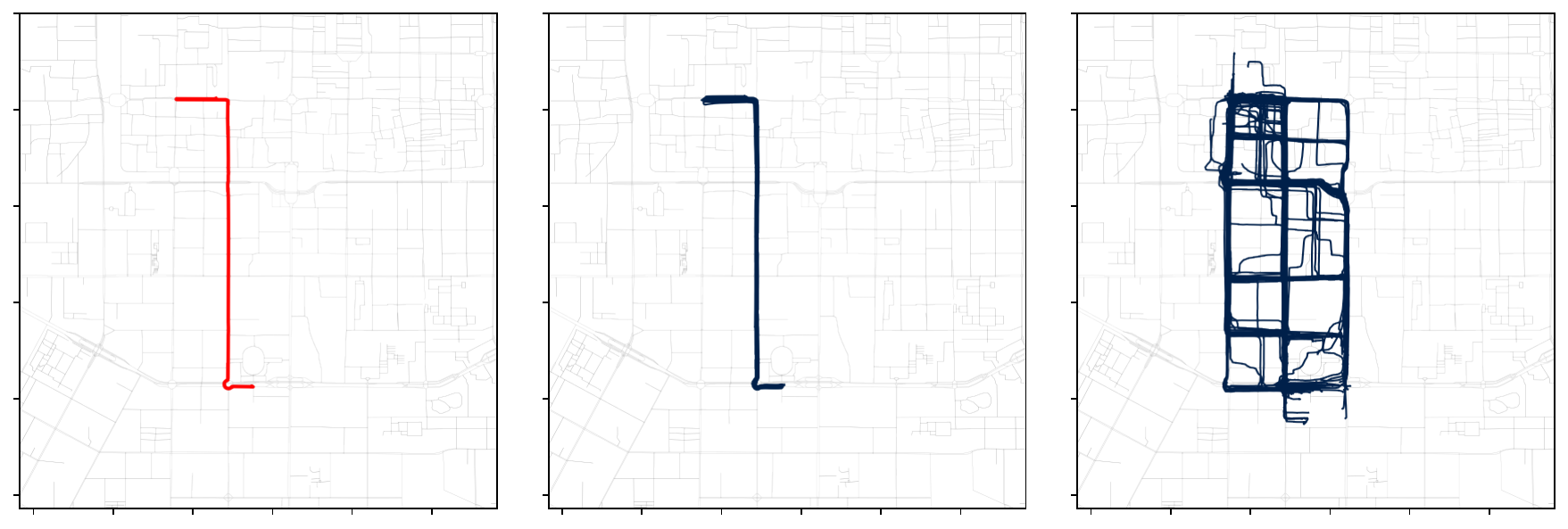}
    }
    \subfigure{
        \includegraphics[width=0.48\linewidth]{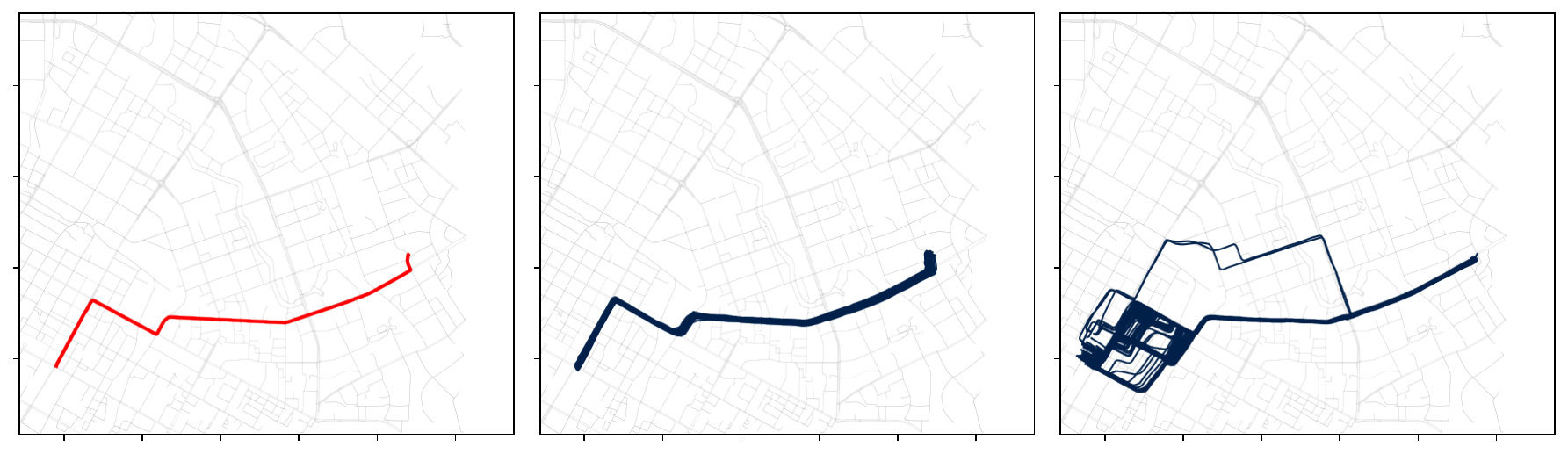}
    }
    \subfigure{
        \includegraphics[width=0.47\linewidth]{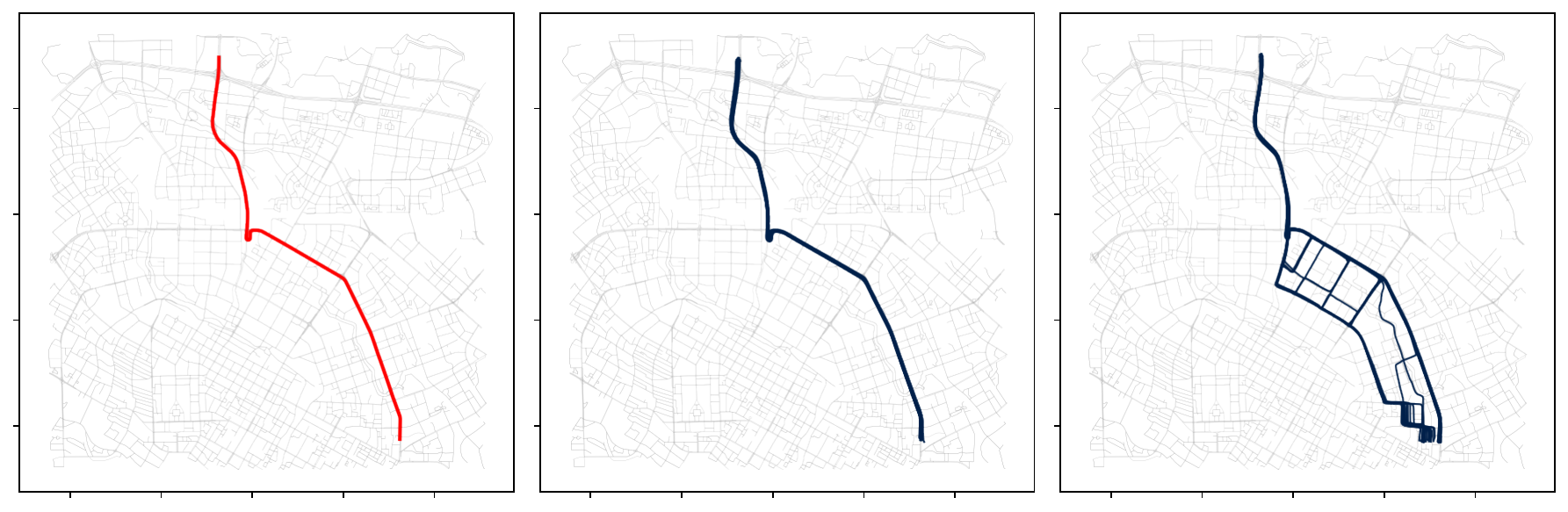}
    }
    \subfigure{
        \includegraphics[width=0.48\linewidth]{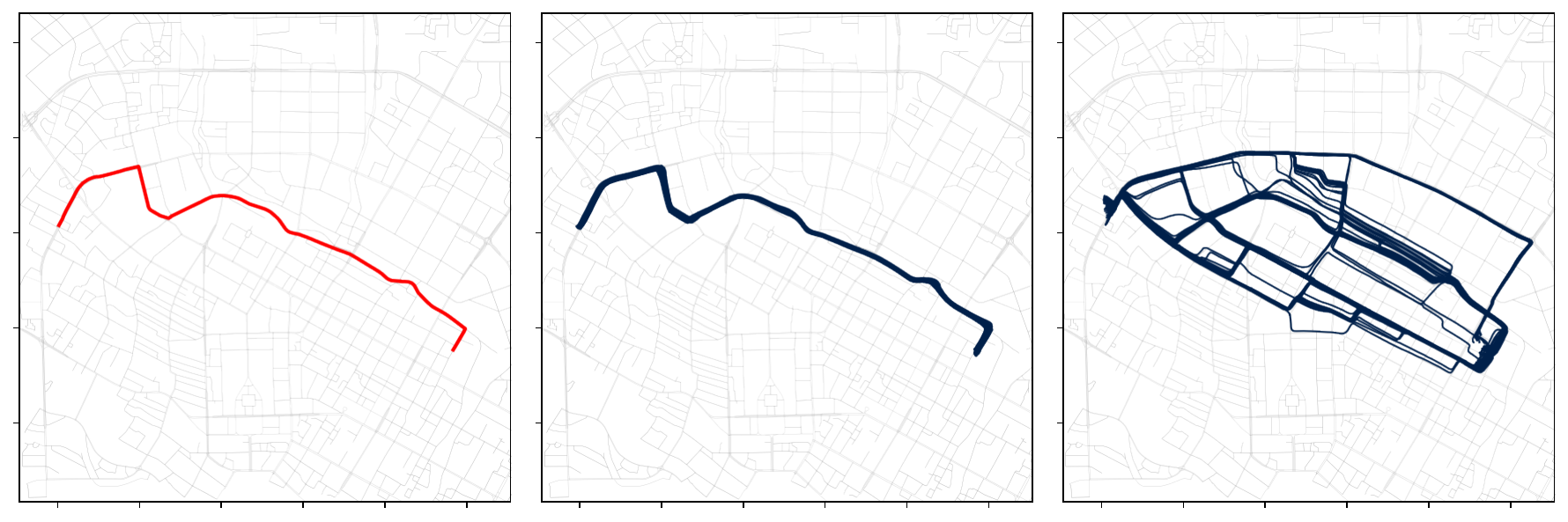}
    }
    \subfigure{
        \includegraphics[width=0.48\linewidth]{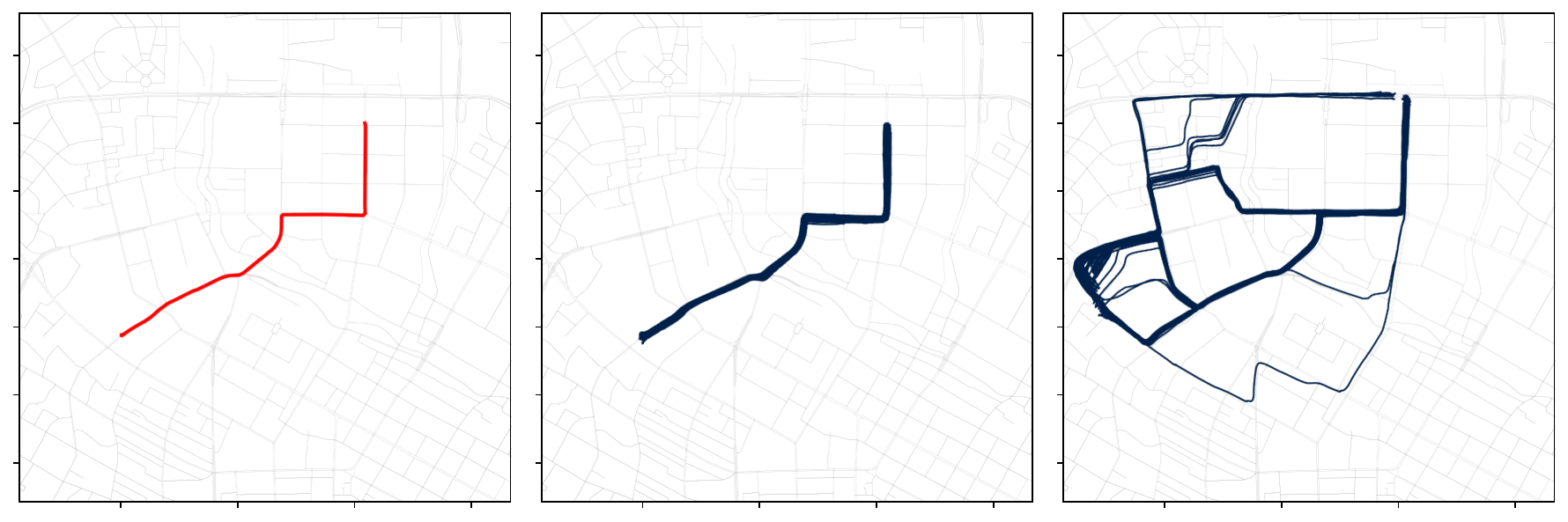}
    }
    \caption{Comparison of controlled trajectory generation in road segment topology constraints.}
    \label{fig:app_control}
\end{figure*}

\subsection{Controllable Trajectories Generation}\label{app:control}
Here, we provide a series of comparisons of the effects of trajectory generation using ControlTraj and ControlTraj for a variety of road segment topology constraint scenarios.
The selected results are displayed in \cref{fig:app_control}.

\subsection{RoadMAE Analysis}\label{app:roadmae}
\begin{figure*}
    \centering
    \subfigure{
        \includegraphics[width=0.48\linewidth]{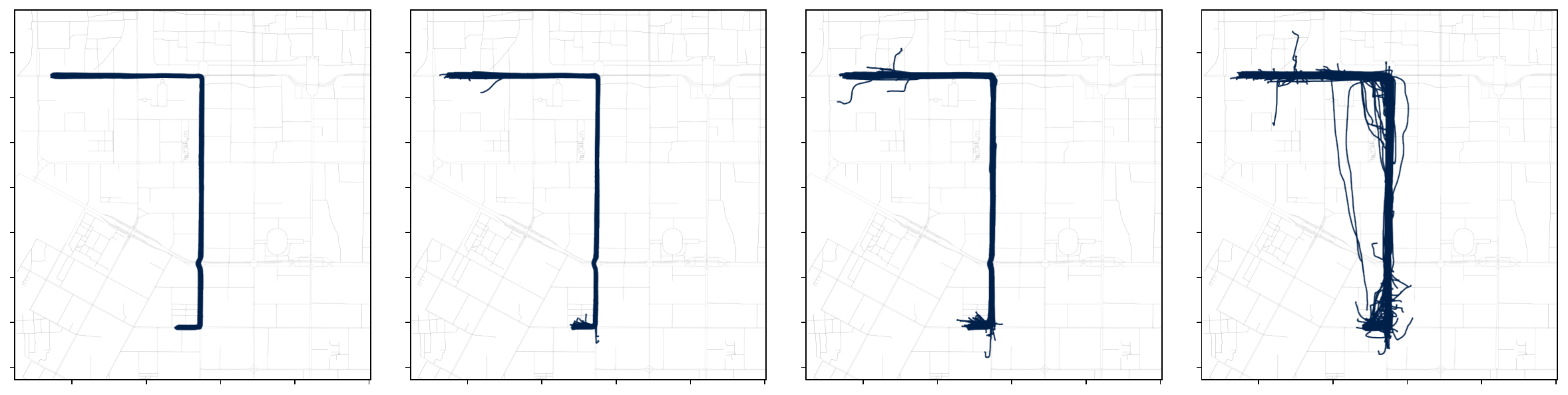}
    }
    \subfigure{
        \includegraphics[width=0.48\linewidth]{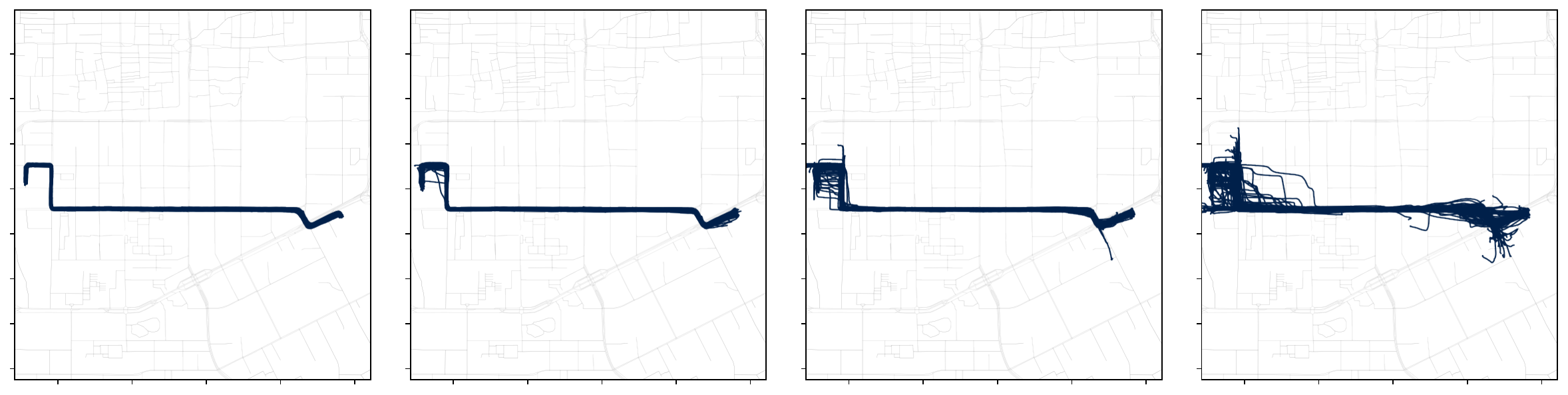}
    }
    \subfigure{
        \includegraphics[width=0.48\linewidth]{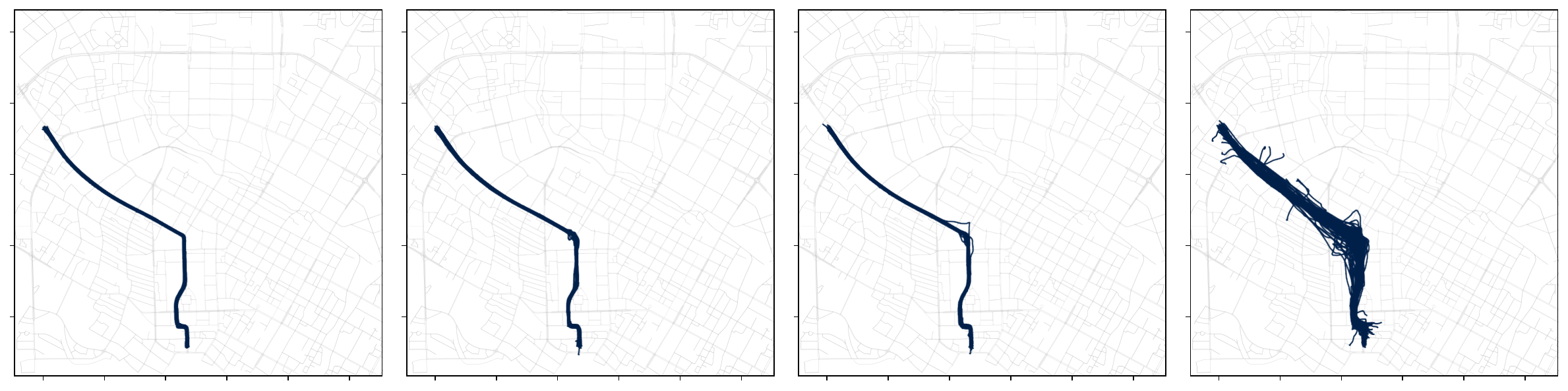}
    }
    \subfigure{
        \includegraphics[width=0.48\linewidth]{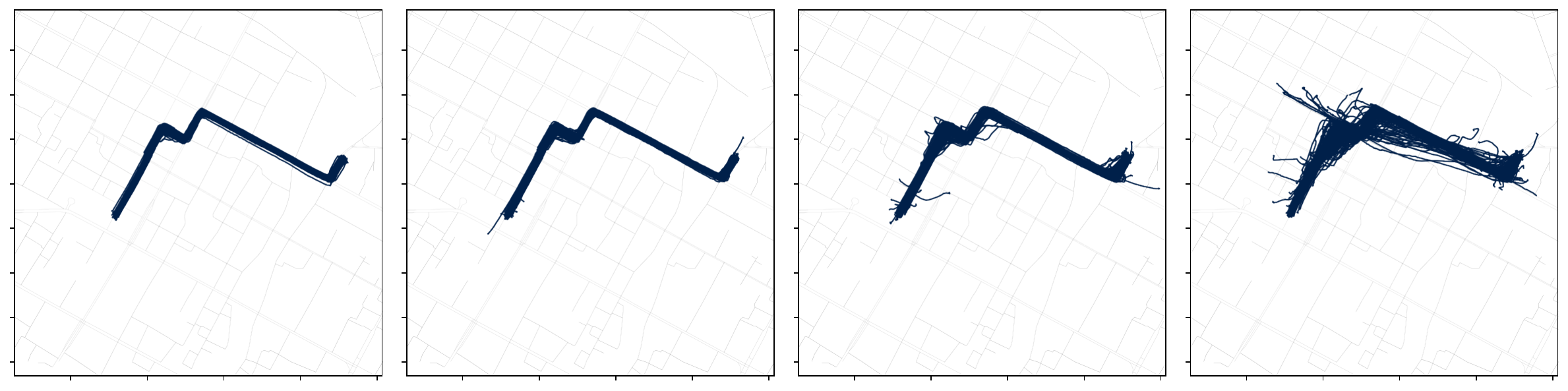}
    }
    \subfigure{
        \includegraphics[width=0.48\linewidth]{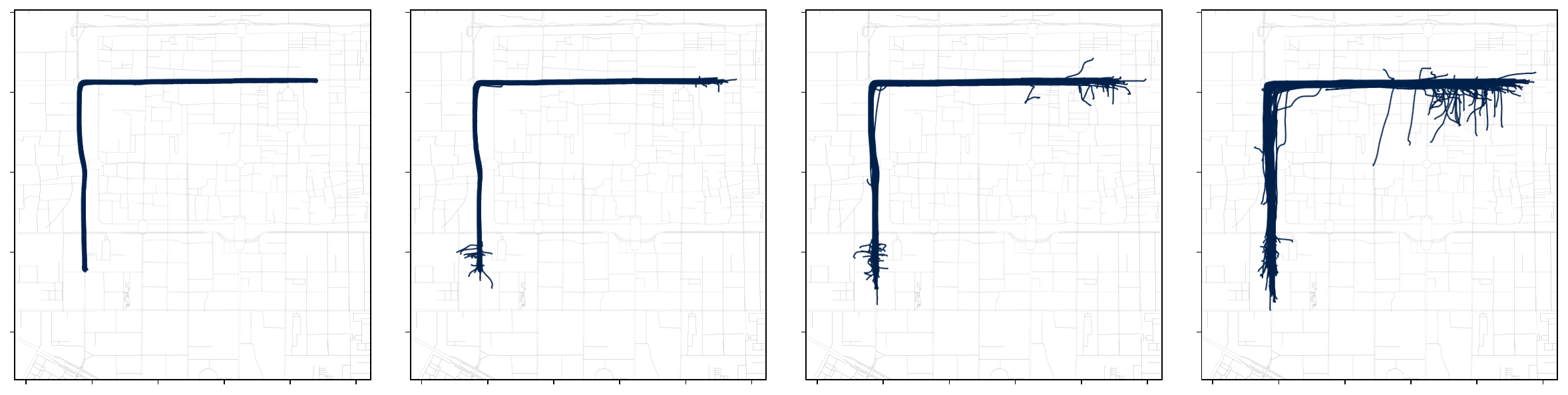}
    }
    \subfigure{
        \includegraphics[width=0.48\linewidth]{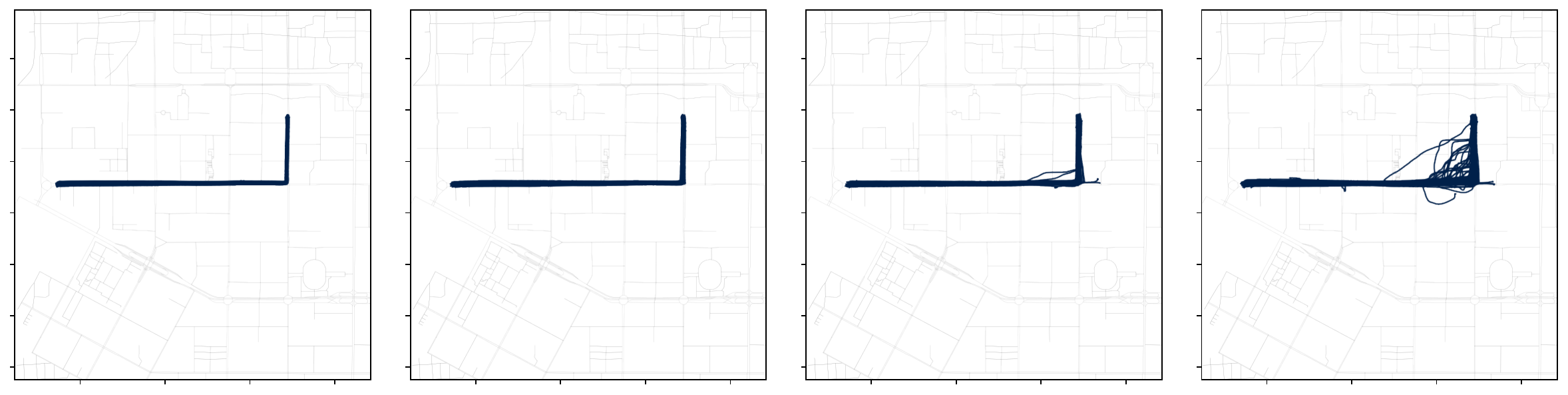}
    }
    \caption{Controllable generation with different mask ratio $\{0, 0.25, 0.5, 0.75 \}$ from left to right.}
    \label{fig:app_roadmae}
\end{figure*}

In this section, we provide more results for RoadMAE trajectory generation at different mask ratios in \cref{fig:app_roadmae}.

\subsection{Generalizability}\label{app:generalizability}
\begin{figure*}[!h]
    \includegraphics[width=0.9\linewidth]{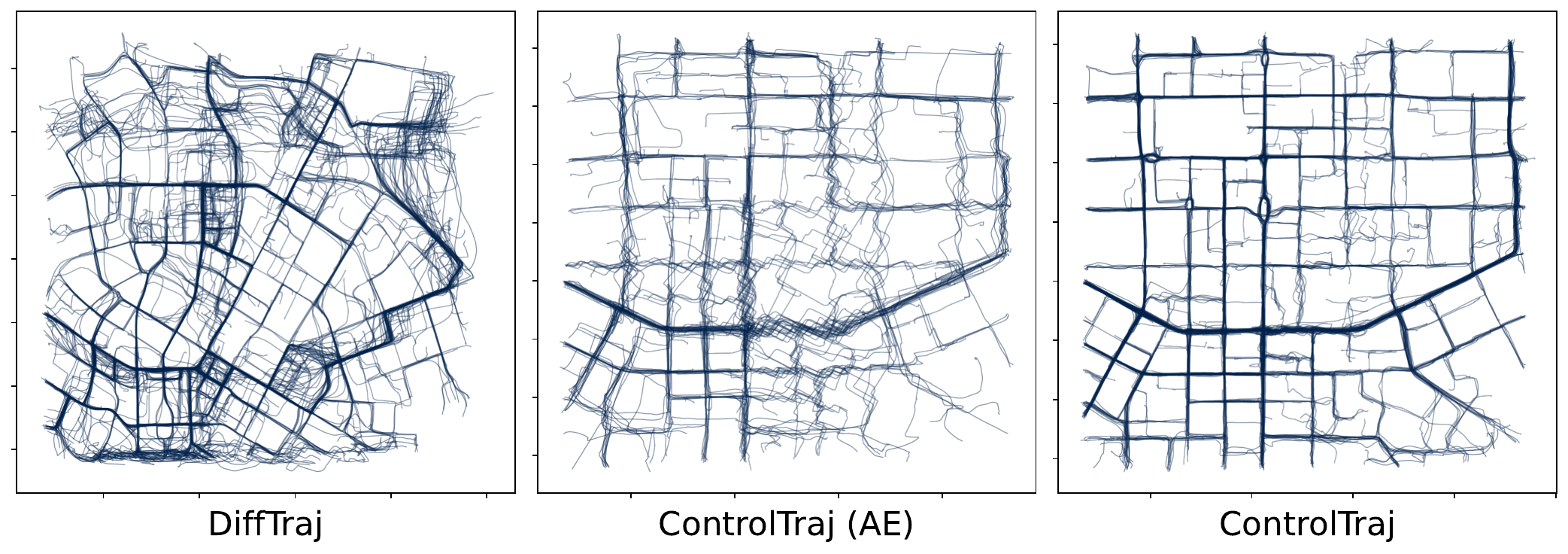}
    \caption{Zero-shot learning for Generalizability (Chengdu $\rightarrow$ Xi'an).}
    \label{fig:app_zero-shot_xa}
\end{figure*}

\begin{figure*}[!h]
    \includegraphics[width=0.9\linewidth]{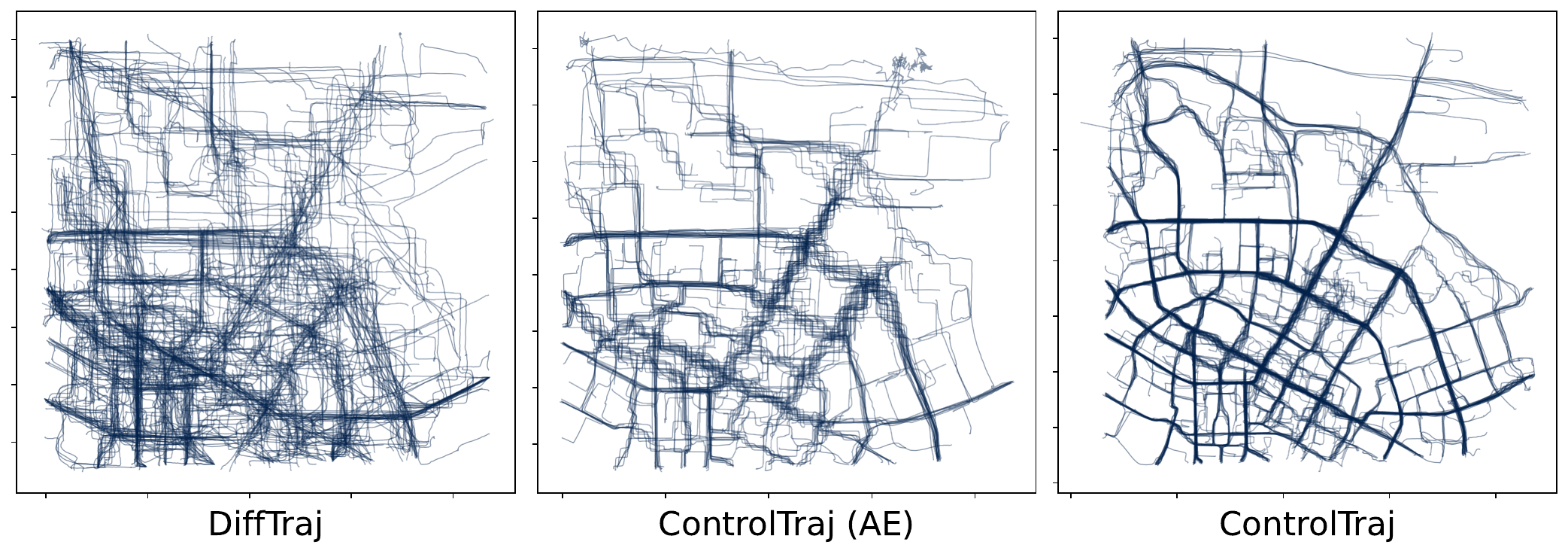}
    \caption{Zero-shot learning for Generalizability (Xi'an $\rightarrow$ Chengdu).}
    \label{fig:app_zero-shot_cd}
\end{figure*}

In this section, we provide visualizations of the results of Chengdu transfer to Xi'an, and Xi'an transfer to Chengdu in a zero-shot learning scenario.
The large view result are shown in \cref{fig:app_zero-shot_xa} and \cref{fig:app_zero-shot_cd}.

\end{appendix}

\end{sloppypar}
\end{document}